\documentclass{article}

\usepackage{microtype}
\usepackage{graphicx}
\usepackage{booktabs}
\usepackage{hyperref}
\usepackage{url}
\usepackage[preprint]{icml2026}

\usepackage{amsmath}
\usepackage{amssymb}
\usepackage{mathtools}
\usepackage{amsthm}
\usepackage{amsfonts}
\usepackage{nicefrac}
\usepackage{xcolor}
\usepackage{multirow}
\usepackage{float}
\usepackage[most]{tcolorbox}
\usepackage{algorithm}
\usepackage{algorithmic}
\usepackage{enumitem}
\usepackage[capitalize,noabbrev]{cleveref}

\tcbset{
    myquote/.style={
        colback=pink!20,
        colframe=pink,
        fonttitle=\bfseries,
        coltitle=black,
        boxrule=0.5mm,
        sharp corners,
        left=1mm,
        right=1mm,
        top=1mm,
        bottom=1mm
    }
}

\newcommand{\answerTODO}[1][]{\textcolor{red}{\bfseries [TODO]}}

\icmltitlerunning{ClawForge: Generating Executable Interactive Benchmarks for Command-Line Agents}

\begin{document}

\twocolumn[

\icmltitle{ \vspace{-9pt}\raisebox{-0.23cm}{
\includegraphics[width=0.05\textwidth]{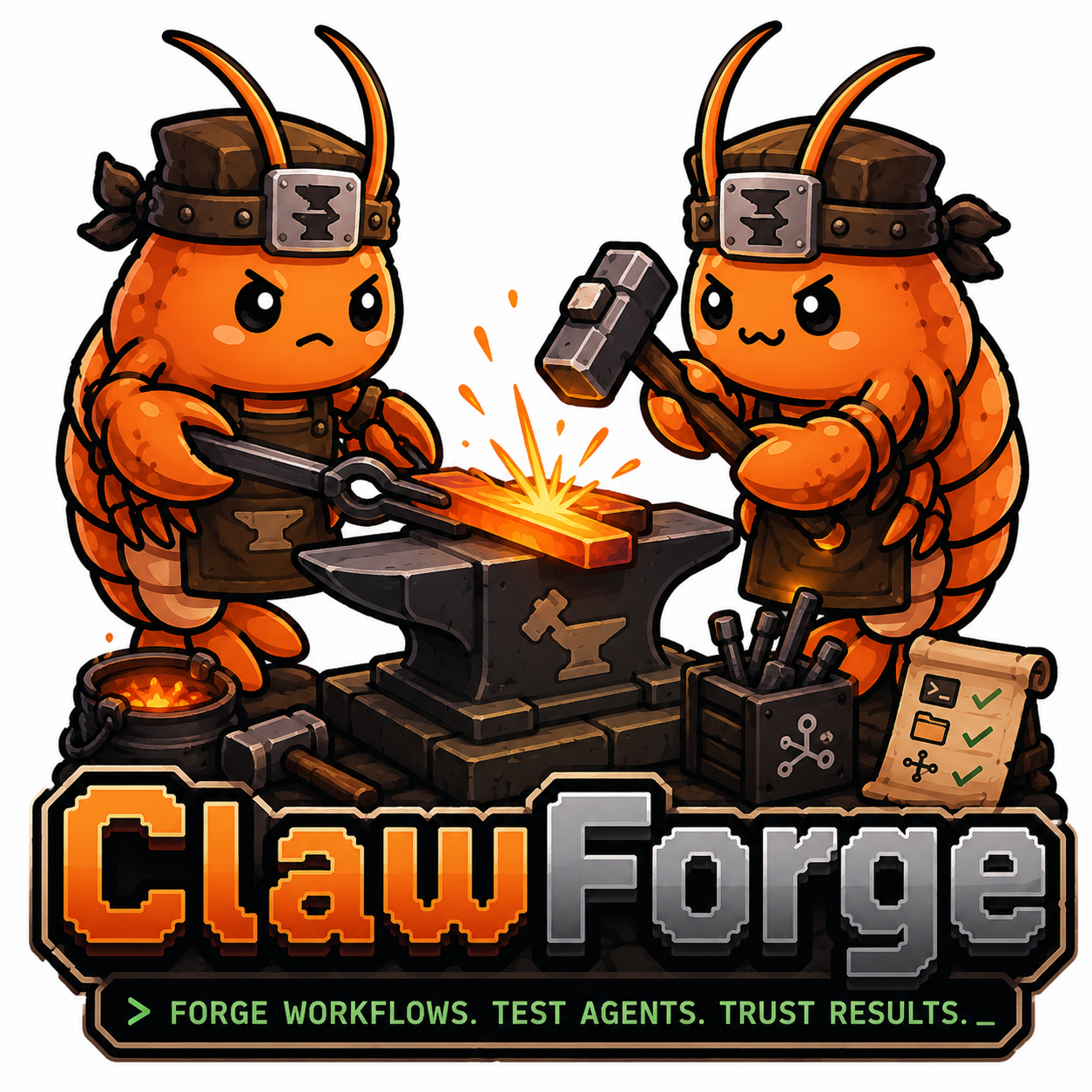}
}
ClawForge: Generating Executable Interactive Benchmarks 
for Command-Line Agents}

\icmlsetsymbol{equal}{*}
\icmlsetsymbol{email}{\Letter}

\begin{icmlauthorlist}
\icmlauthor{Yuxiang Lai}{aff1,equal}
\icmlauthor{Peng Xia}{aff1,equal}
\icmlauthor{Haonian Ji}{aff1}
\icmlauthor{Kaiwen Xiong}{aff1}
\icmlauthor{Kaide Zeng}{aff1}
\icmlauthor{Jiaqi Liu}{aff1}
\icmlauthor{Fang Wu}{aff2}
\icmlauthor{Jike Zhong}{aff4}
\icmlauthor{Zeyu Zheng}{aff3}
\icmlauthor{Cihang Xie}{aff5}
\icmlauthor{Huaxiu Yao}{aff1}
\end{icmlauthorlist}

\icmlaffiliation{aff1}{University of North Carolina at Chapel Hill}
\icmlaffiliation{aff2}{Stanford University}
\icmlaffiliation{aff3}{University of California, Berkeley}
\icmlaffiliation{aff4}{University of Southern California}
\icmlaffiliation{aff5}{University of California, Santa Cruz}

\icmlcorrespondingauthor{Huaxiu Yao}{huaxiu@cs.unc.edu}

\vskip 0.3in
]

\printAffiliationsAndNotice{\icmlEqualContribution: \textless{}yxlai@unc.edu, pxia@cs.unc.edu\textgreater{}}

\begin{abstract}
Interactive agent benchmarks face a tension between scalable construction and realistic workflow evaluation. Hand-authored tasks are expensive to extend and revise, while static prompt evaluation misses failures that only appear when agents operate over persistent state. Existing interactive benchmarks have advanced agent evaluation significantly, but most initialize tasks from clean state and do not systematically test how agents handle pre-existing partial, stale, or conflicting artifacts. We present \textbf{ClawForge}, a generator-backed benchmark framework for executable command-line workflows under state conflict. The framework compiles scenario templates, grounded slots, initialized state, reference trajectories, and validators into reproducible task specifications, and evaluates agents step by step over persistent workflow surfaces using normalized end state and observable side effects rather than exact trajectory matching. We instantiate this framework as the ClawForge-Bench (17 scenarios, 6 ability categories). Results across seven frontier models show that the best model reaches only 45.3\% strict accuracy, wrong-state replacement remains below 17\% for all models, and the widest model separation (17\% to 90\%) is driven by whether agents inspect existing state before acting. Partial-credit and step-efficiency analyses further reveal that many failures are near-miss closures rather than early breakdowns, and that models exhibit qualitatively different failure styles under state conflict. 
Code and Data: \url{https://github.com/aiming-lab/ClawForge}
\end{abstract}

\section{Introduction}
\label{sec:introduction}

As Large Language Model (LLM) agents~\cite{yao2022react,shinn2023reflexion} move from one-shot prompting to persistent software workflows, benchmark construction becomes a bottleneck in its own right. A realistic workflow task is not just an instruction paired with an answer: it requires initialized state, valid and invalid pre-existing artifacts, expected side effects, and executable validation logic. These ingredients must also be reproducible and auditable, because small changes in state can turn an apparently equivalent task into a different decision problem. Hand-authored interactive tasks are therefore expensive to scale and difficult to revise once coverage or fairness issues emerge~\citep{zheng2023judging,chang2024survey}.

This problem is especially acute for command-line agents~\cite{openclaw,merrill2026terminal,hermes-agent}. Real workflows require agents to inspect task boards, inbox threads, calendars, files, runtime configuration, weather state, and messaging surfaces before deciding whether to add, preserve, repair, or replace state. Many failures only appear after execution: an agent may create a duplicate task, leave stale state unresolved, choose the wrong branch under conflicting evidence, or stop before the final side effect is produced. Existing interactive benchmarks have advanced agent evaluation significantly~\citep{qin2023toolllm,liu2023agentbench,zhou2023webarena}, but most initialize tasks from clean state and do not systematically test how agents handle pre-existing partial, stale, or conflicting artifacts~\cite{ye2026claw,ji2026clawarena,li2026clawsbench}. This means a central part of workflow competence remains unmeasured: correctness that depends on the final state of a live environment, not only on a plausible textual answer or command sequence.

\begin{figure*}[t]
    \centering
    \includegraphics[width=1\linewidth]{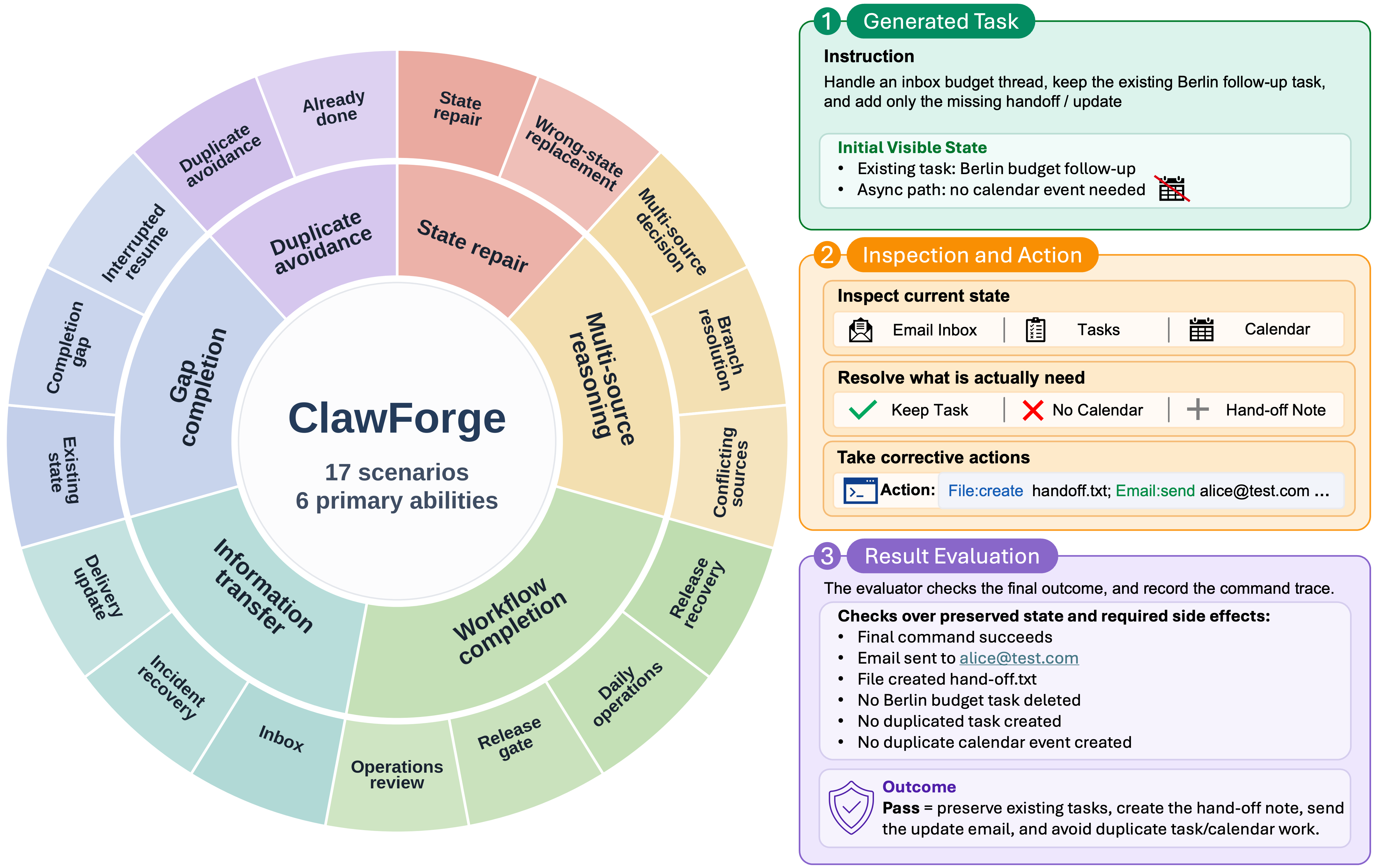}
    \caption{ClawForge-Bench benchmark coverage. Inner: 6 primary ability categories. Outer: 17 scenario families within each category.}
    \label{fig:benchmark_composition}
\end{figure*}

To address these challenges, we present \textbf{ClawForge}, a generator-backed benchmark framework for executable state-conflict workflows. ClawForge compiles scenario templates, grounded variables, seeded environment state, reference trajectories, validators, and metadata into executable task specifications that can be regenerated, audited, and systematically extended. The execution environment then evaluates agents step by step through a CLI-style interface, scoring normalized workflow state and observable side effects rather than exact command imitation. Automatic generation is therefore not merely a scalable way to produce more tasks, but a mechanism for maintaining reproducibility, extensibility, and evaluation consistency in interactive benchmarks. 
We instantiate this framework as \emph{ClawForge-Bench}, a suite of \textbf{17 scenarios} organized into six ability categories (\Cref{fig:benchmark_composition}) that begin from partial, stale, or conflicting workflow state and target realistic failure modes, including duplicate-aware completion, stale-state repair, wrong-state replacement, multi-source branch resolution, and full workflow closure.

In summary, our primary contribution is ClawForge, a generator-backed benchmark framework for evaluating command-line agents under state-conflict workflows through automated task construction, stateful execution, and result-first evaluation. On ClawForge-Bench (17 scenarios spanning 6 ability categories), evaluations across seven frontier models show that the benchmark remains far from saturated: the best model achieves only 45.3\% strict accuracy, wrong-state replacement stays below 17\% for all models, and Interrupted Workflow Resume exhibits the largest model gap (17\%--90\%), largely depending on whether agents inspect existing state before acting. Partial-credit and step-efficiency analyses further reveal that many failures arise from near-miss workflow closure rather than early breakdowns, and that models exhibit qualitatively different failure behaviors under state conflict.

\section{Task Generation and Execution}
\label{sec:method}

ClawForge is a generator-backed benchmark system rather than a prompt collection. The central design principle is that each generated task is an executable specification: the instruction, initialized state, reference trajectory, validators, and metadata are produced together and then evaluated through the same stateful runtime. This section describes how tasks are generated (\S\ref{sec:generation}) and executed (\S\ref{sec:environment}); \Cref{sec:setup} defines the evaluation protocol.

\subsection{Automated Benchmark Generation}
\label{sec:generation}

\begin{figure*}[t]
    \centering
    \includegraphics[width=\linewidth]{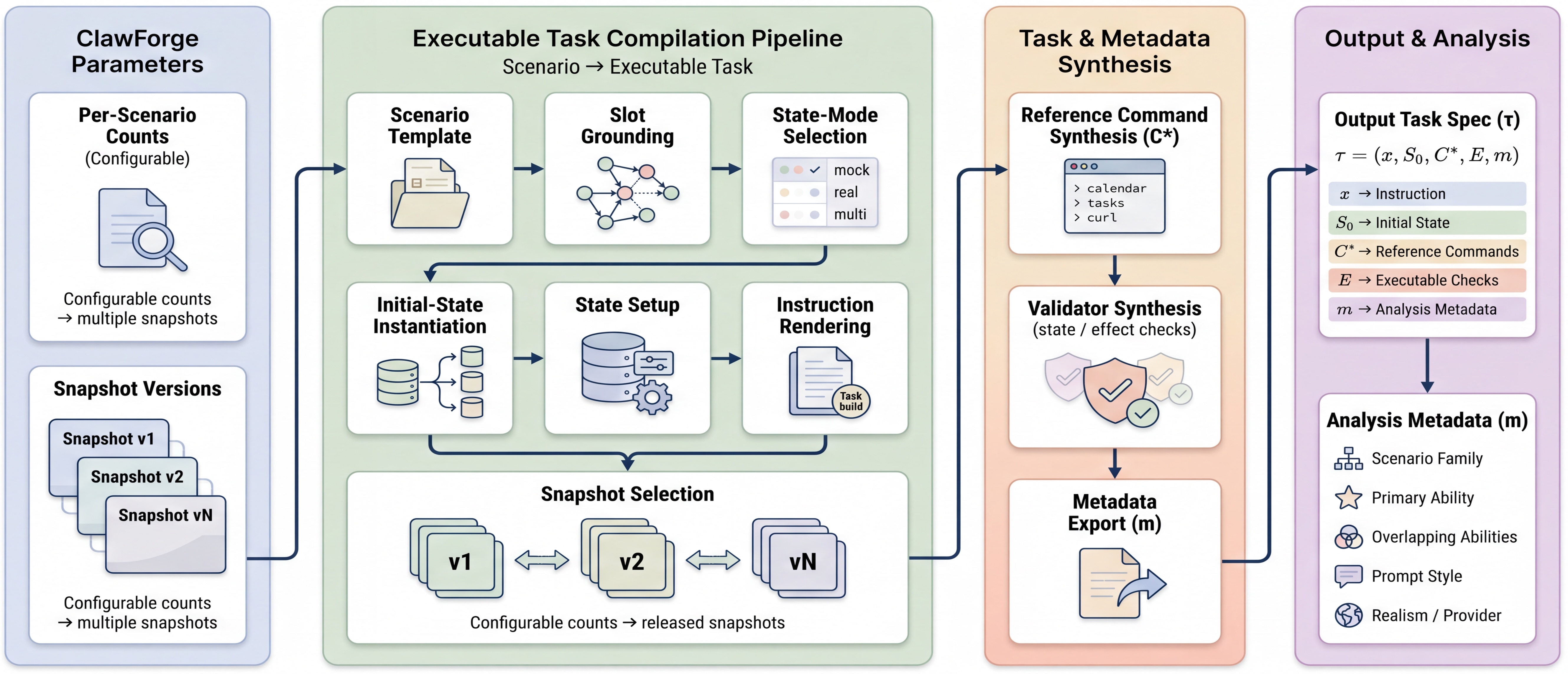}
    \caption{Automated benchmark generation pipeline. Scenario templates are compiled into executable task specifications $\tau = (x, S_0, C^\star, \mathcal{E}, m)$ through slot grounding, state initialization, instruction rendering, reference command synthesis, and validator generation.}
    \label{fig:generation_pipeline}
\end{figure*}

Each task is generated from a scenario template together with grounded slots such as city, timezone, recipient, topic, due date, or calendar start time. This yields surface diversity while ensuring that every generated task maps back to a known scenario and can be evaluated by the same validators. As shown in \Cref{fig:generation_pipeline}, the generation pipeline proceeds in stages: the scenario template is first grounded with concrete slot values, then a state-mode (mock or real) is selected and the initial environment state is instantiated accordingly, followed by instruction rendering. The pipeline then synthesizes reference commands $C^\star$ by executing the intended workflow against the initialized state, generates validators that check required state transitions and side effects, and exports structured metadata (scenario family, primary ability, prompt style). The final output is a complete executable task object.

A generated task is an executable specification with five coupled components:
\begin{equation}
\tau = (x, S_0, C^\star, \mathcal{E}, m),
\end{equation}
where $x$ is the instruction, $S_0$ the initialized state, $C^\star$ a reference command trajectory, $\mathcal{E}$ the executable checks, and $m$ the structured metadata. Each $\tau$ is instantiated from a scenario family $\sigma$, a grounded slot assignment $z$, and a prompt policy $p$. The key design choice is that all five components are generated together from the same scenario specification, so the benchmark object is an executable workflow task rather than a prompt paired with an offline answer key.

Because the generator is parameterized by scenario family, different scenarios can encode fundamentally different decision structures rather than only surface wording variations. Some are gap-completion tasks, some require explicit repair or replacement of stale state, and others require branch selection across several information sources. This structural diversity lets the benchmark separate state conflict, closure efficiency, and multi-source decision failures rather than measuring only surface command following (\Cref{sec:experiment}). Once generated, each task is executed through the interactive environment described next.

\subsection{Interactive Environment}
\label{sec:environment}

\begin{figure*}[t]
    \centering
    \includegraphics[width=\linewidth]{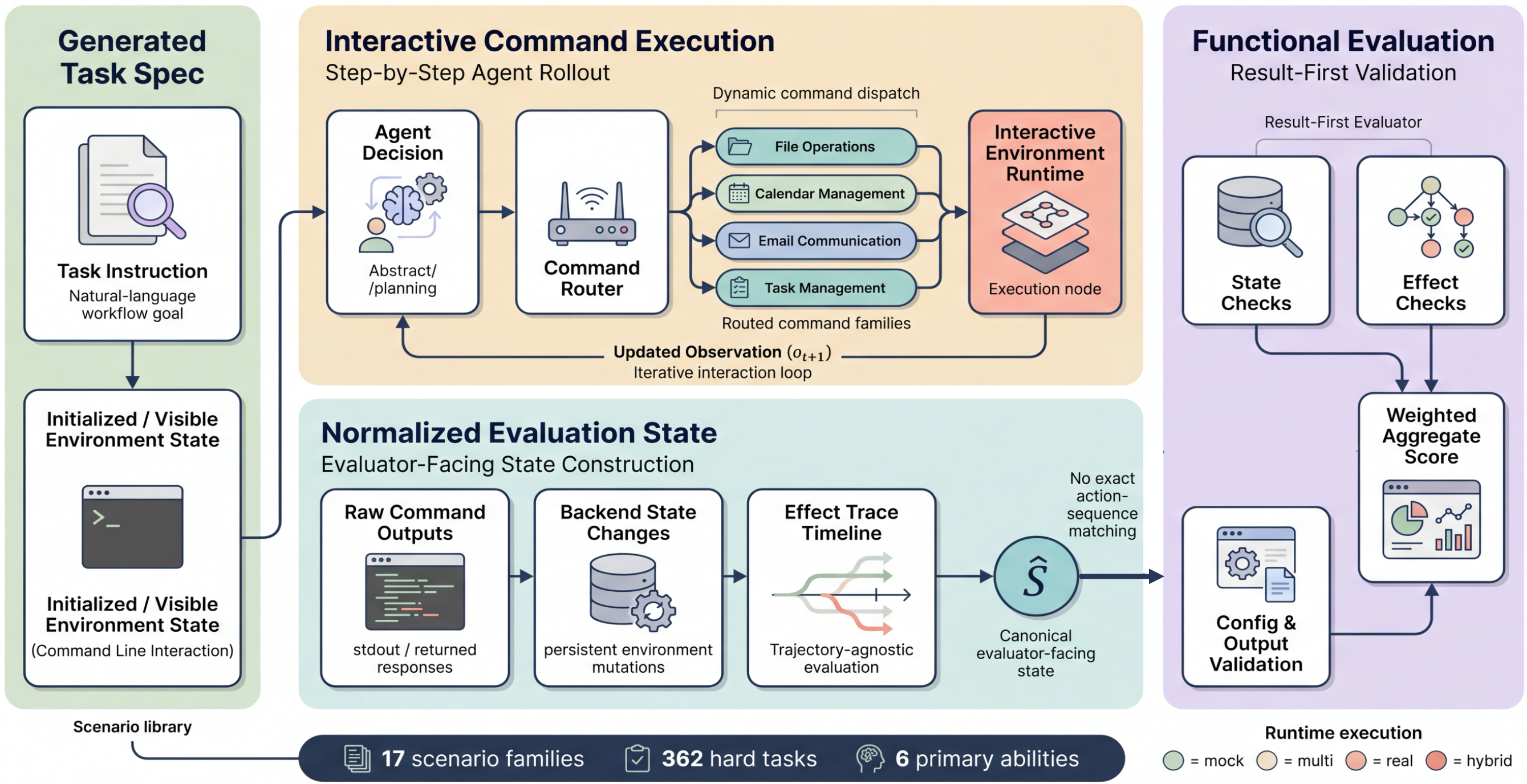}
    \caption{Interactive execution and evaluation loop. Agents emit commands step by step; the environment executes them, records state changes and effect traces, and merges everything into a normalized evaluation state $\hat{S}$ for result-first scoring.}
    \label{fig:loop}
\end{figure*}

Generated tasks are executed in a stateful environment rather than graded as static prompts. ClawForge exposes a CLI-style interface over workflow surfaces such as tasks, calendar, email, messaging, files, runtime configuration, weather, and recurring checks. An episode begins with a natural-language instruction. At each step, the agent emits one command, the environment executes it, and the next observation is built from the resulting outputs and updated state. This execution-and-evaluation loop is summarized in \Cref{fig:loop}, which makes explicit that execution and result-first grading happen inside one coupled episode rather than as separate offline stages.

This loop is intentionally stateful: many failures only become visible after execution. An agent may recreate an existing object, leave stale state untouched, update the wrong entity, or stop before the final side effect is produced. ClawForge therefore treats execution state as a first-class object in both rollout and evaluation (see Appendix~\ref{sec:appendix_exec_details} for implementation details).

\section{Evaluation Protocol}
\label{sec:setup}

\begin{algorithm}[t]
\caption{ClawForge episode rollout}
\label{alg:rollout_main}
\begin{algorithmic}[1]
\STATE Load task $T$ and copy base state $S_0$ into an isolated state directory
\STATE Apply task-specific state overrides
\STATE Initialize routed backend $B$ and evaluator $E$
\STATE Initialize observation $o_0 \leftarrow$ instruction, config, gateway status, command hints
\FOR{$t = 1, \dots, H$}
    \IF{agent requests stop or emits \texttt{DONE}}
        \STATE break
    \ENDIF
    \STATE Agent emits one command $a_t$
    \STATE Execute $a_t$ through routed backend $B$
    \STATE Record stdout/stderr, exit code, command metadata, and inferred effects
    \STATE Construct next observation $o_t$
    \IF{$a_t \in \{\texttt{done}, \texttt{exit}, \texttt{quit}\}$ or $t = H$ or a rollout stopping rule triggers}
        \STATE break
    \ENDIF
\ENDFOR
\STATE Build normalized evaluation state $\hat{S}$ from command history, effects, latest outputs, config, and merged backend state
\STATE Return final result $E(\hat{S}, T)$
\end{algorithmic}
\end{algorithm}

ClawForge evaluates tasks functionally rather than by comparing surface-form trajectories. The evaluator operates over normalized execution state and explicit effect traces, so multiple trajectories can pass as long as they produce the required state transitions and observable side effects. Because tasks are generated together with their validators, evaluation stays tied to the same scenario semantics that produced the task. This result-first design lets us distinguish between early breakdowns and structured near misses, such as preserving the wrong object, omitting one final side effect, or stopping after a partially correct repair.

\subsection{Episode Rollout}
\label{sec:rollout}

Each episode is instantiated from a generated task containing an instruction, initialized state, reference trajectory, executable checks, and structured metadata. On reset, the environment materializes task state, applies any task-specific state overrides, initializes the selected backend, and builds the evaluator. The agent receives the instruction once and then emits exactly one command at each subsequent step. The environment executes that command, records state-changing effects, and returns the next observation. Algorithm~\ref{alg:rollout_main} summarizes this interaction. When a rollout ends, the environment builds a normalized state $\hat{S}$ from the command history, accumulated effect traces, the latest process outputs, and the merged backend state. The evaluator then applies task-defined checks over $\hat{S}$ rather than comparing against one exact command sequence. The default protocol uses \texttt{multi} mode, which routes commands to their corresponding workflow surfaces, with a configurable maximum step limit (see Appendix~\ref{sec:appendix_exec_details} for additional runtime details).

\subsection{Scoring}
\label{sec:scoring}

We report two complementary metrics. Let $\mathcal{D}$ denote the evaluation set. \emph{Strict full-pass accuracy} counts the fraction of tasks where every required check passes:
\begin{equation}
\mathrm{Acc}(\mathcal{D}) = \frac{1}{|\mathcal{D}|}\sum_{\tau \in \mathcal{D}} \mathbf{1}[\tau \text{ passes all required checks}].
\end{equation}
\emph{Partial-credit score} measures how much of each workflow was completed correctly. If a rollout has checks $\{c_i\}_{i=1}^n$, each with score $s_i \in [0,1]$ and weight $w_i > 0$, the per-task score is
\begin{equation}
\mathrm{Score}(\hat{S}) = \frac{\sum_{i=1}^{n} w_i s_i}{\sum_{i=1}^{n} w_i},
\end{equation}
and we report the average over $\mathcal{D}$. The two metrics use the same evaluator but summarize it differently: strict accuracy rewards complete closure, while partial credit distinguishes near-miss failures from early breakdowns. When agents are served by external providers, ClawForge also records provider-side failures and provider-impacted tasks as supplementary records.

\section{Experiments}
\label{sec:experiment}

We evaluate the ClawForge-Bench (17 scenarios, 362 tasks, two prompt styles) across seven frontier models. Our experiments address the following questions: (1)~Is the benchmark challenging and discriminative across strong models? (2)~Which ability categories expose the largest capability gaps? (3)~Do models differ in step efficiency, and does extra interaction translate into better closure? (4)~Can partial-credit analysis separate near-miss failures from early stops?

\subsection{Experimental Setup}

All experiments follow the protocol in \Cref{sec:setup} using \texttt{multi} mode, full interaction history, and a 25-step budget. We evaluate seven frontier models: Kimi-K2.5~\citep{team2026kimi}, GPT-5.2~\citep{openai2025gpt52}, OpenAI o3~\citep{openai2025o3}, GPT-5.3-Codex~\citep{openai2025gpt53}, GPT-5.4~\citep{openai2026gpt54}, Claude Sonnet 4.6~\citep{claude2026sonnet46}, and Claude Opus 4.6~\citep{claude2026opus46}. We report strict full-pass accuracy and partial-credit score as defined in \Cref{sec:scoring}. For Claude models, step counts in \Cref{tab:length_results} are aggregated from partially overlapping trace subsets rather than a single uniform run; consequently, step-efficiency comparisons with other models should be interpreted cautiously (see Appendix~\ref{sec:appendix_data_provenance}).

\subsection{Benchmark Results}

The generated ClawForge-Bench snapshot separates abilities that matter in stateful execution. \Cref{tab:main_results,tab:ability_results,tab:length_results,tab:ability_score_results} highlight three benchmark findings. First, aggregate performance remains far from saturation while still separating strong frontier models. Second, the hardest settings are not simply long workflows, but scenarios involving stale state, incorrect replacement, and incomplete workflow closure. Third, partial-credit and step-efficiency analyses expose meaningful behavioral differences---including near-miss completion and inefficient execution---that strict pass/fail accuracy alone would largely obscure.

\subsubsection{Overall Benchmark Separation}

\Cref{tab:main_results} shows that the benchmark is neither saturated nor dominated by one model family: the best strict accuracy is 45.3\% (Claude Opus 4.6) and the weakest is 35.9\% (Kimi-K2.5). Strict accuracy and partial credit rank models differently: GPT-5.3-Codex has the highest partial credit (0.8238) but ranks fourth on strict accuracy, while Claude Opus 4.6 leads strict accuracy but has lower partial credit (0.7868). This mismatch indicates that the suite distinguishes models that steadily accumulate correct intermediate state from models that convert a smaller subset of trajectories into complete workflow closure.

\begin{table*}[t]
    \centering
    \caption{\textbf{Main results} (strict accuracy and partial score). All entries use \texttt{multi} mode with full interaction history and a 25-step budget.}
    \label{tab:main_results}
    \setlength{\tabcolsep}{8pt}
    \begin{tabular}{l|cc|c}
        \toprule
        Model & Strict Acc. & Partial Credit & Rank \\
        \midrule
        Claude Opus 4.6~\citep{claude2026opus46} & \textbf{45.3\%} & 0.7868 & \textbf{1} \\
    Claude Sonnet 4.6~\citep{claude2026sonnet46} & 42.3\% & 0.7473 & 2 \\
        GPT-5.4~\citep{openai2026gpt54} & 40.9\% & 0.8078 & 3 \\
        GPT-5.3-Codex~\citep{openai2025gpt53} & 40.6\% & \textbf{0.8238} & 4 \\
        GPT-o3~\citep{openai2025o3} & 39.5\% & 0.8071 & 5 \\
        GPT-5.2~\citep{openai2025gpt52} & 38.7\% & 0.7883 & 6 \\
        Kimi-K2.5~\citep{team2026kimi} & 35.9\% & 0.7404 & 7 \\
        \bottomrule
    \end{tabular}
\end{table*}

\begin{table*}[t]
    \centering
    \caption{\textbf{Strict accuracy (\%) by primary ability.} Each row is a disjoint partition of the task set.}
    \label{tab:ability_results}
    \begin{tabular}{l|c|cccc|cc}
        \toprule
        \multirow{2}{*}{Primary Ability} & Kimi & \multicolumn{4}{c|}{GPT} & \multicolumn{2}{c}{Claude} \\
        \cmidrule(lr){2-8}
         & K2.5 & 5.2 & o3 & 5.3-Codex & 5.4 & Sonnet 4.6 & Opus 4.6 \\
        \midrule
        duplicate avoidance & 95.0\% & 95.0\% & 95.0\% & 95.0\% & \textbf{100.0\%} & 100.0\% & 95.0\% \\
        gap completion & 26.9\% & 42.3\% & 50.0\% & 51.9\% & 53.8\% & \textbf{78.8\%} & \textbf{78.8\%} \\
        information transfer & 71.9\% & 62.5\% & 50.0\% & 71.9\% & \textbf{81.2\%} & 59.4\% & 59.4\% \\
        multi-source reasoning & 31.4\% & 31.4\% & 25.6\% & 37.2\% & \textbf{40.7}\% & 39.5\% & 37.2\% \\
        state repair & 18.5\% & 26.1\% & \textbf{33.7\%} & 13.0\% & 12.0\% & 17.4\% & 29.3\% \\
        workflow completion & 37.5\% & 35.0\% & 36.2\% & \textbf{42.5\%} & 35.0\% & 28.8\% & 32.5\% \\
        \bottomrule
    \end{tabular}%
\end{table*}

\begin{table*}[t]
    \centering
    \caption{\textbf{Partial-credit score by primary ability.} Higher scores indicate more complete workflow execution even when evaluation fails.}
    \label{tab:ability_score_results}
    \begin{tabular}{l|c|cccc|cc}
        \toprule
        \multirow{2}{*}{Primary Ability} & Kimi & \multicolumn{4}{c|}{GPT} & \multicolumn{2}{c}{Claude} \\
        \cmidrule(lr){2-8}
         & K2.5 & 5.2 & o3 & 5.3-Codex & 5.4 & Sonnet 4.6 & Opus 4.6 \\
        \midrule
        duplicate avoidance & 0.9930 & 0.9930 & 0.9790 & 0.9930 & \textbf{1.0000} & \textbf{1.0000} & 0.9930 \\
        gap completion & 0.7681 & 0.8813 & 0.7898 & 0.8731 & 0.8735 & \textbf{0.8940} & 0.8810 \\
        information transfer & 0.9300 & 0.9059 & 0.8897 & 0.9259 & \textbf{0.9572} & 0.9060 & 0.9140 \\
        multi-source reasoning & 0.8303 & 0.8020 & 0.7536 & 0.8277 & 0.8206 & 0.7910 & \textbf{0.8670} \\
        state repair & 0.4297 & 0.5886 & \textbf{0.6757} & 0.6755 & 0.6171 & 0.4130 & 0.4980 \\
        workflow completion & 0.8435 & 0.8431 & 0.8431 & \textbf{0.8719} & 0.8605 & 0.8630 & 0.8690 \\
        \bottomrule
    \end{tabular}%
    \vspace{-5pt}
\end{table*}

\begin{table*}[t]
    \centering
\caption{\textbf{Accuracy (\%) and average executed steps by reference-trajectory length.} Format: Acc / Steps. Shorter reference trajectories indicate structurally simpler workflows.}
    \label{tab:length_results}
    \setlength{\tabcolsep}{3pt}
    \begin{tabular}{lc|c|cccc|cc}
        \toprule
        \multirow{2}{*}{GT} & \multirow{2}{*}{Num} & Kimi & \multicolumn{4}{c|}{GPT} & \multicolumn{2}{c}{Claude} \\
        \cmidrule(lr){3-9}
        & & K2.5 & 5.2 & o3 & 5.3-Codex & 5.4 & Sonnet 4.6 & Opus 4.6 \\
        \midrule
        $5$ steps & 77 & 36.4 / 7.8 & 37.7 / 7.3 & 40.3 / 8.4 & 48.1 / 6.2 & 44.2 / 6.3 & 40.3 / 8.1 & 40.3 / 9.6 \\
        $6$ steps & 115 & 41.7 / 10.3 & 40.0 / 10.7 & 49.6 / 10.3 & 42.6 / 6.6 & 39.1 / 8.1 & 38.3 / 14.5 & 46.1 / 16.6 \\
        $7$ steps & 45 & 53.3 / 11.7 & 53.3 / 13.7 & 51.1 / 12.4 & 28.9 / 8.2 & 33.3 / 9.6 & 31.1 / 20.7 & 37.8 / 19.9 \\
        $8$ steps & 53 & 22.6 / 10.8 & 41.5 / 10.7 & 39.6 / 11.9 & 43.4 / 8.5 & 50.9 / 8.8 & 69.8 / 17.7 & 67.9 / 15.8 \\
        $9$ steps & 72 & 25.0 / 10.3 & 26.4 / 12.0 & 15.3 / 9.4 & 34.7 / 8.1 & 37.5 / 8.5 & 37.5 / 14.8 & 37.5 / 14.9 \\
        \bottomrule
    \end{tabular}%
    \vspace{-10pt}
\end{table*}

\subsubsection{Ability-Level Analysis}

The clearest capability gap appears when the environment already contains partially wrong state (\Cref{tab:ability_results}). State repair remains below 34\% for all models, and Wrong-State Replacement is the hardest shared slice: even the best model reaches only 17.4\%, while several remain at 0.0\%. These failures are qualitatively different from ordinary completion errors, as the model must identify stale state, preserve valid parts, retire or replace invalid parts, and still finish downstream follow-through.

\Cref{tab:ability_score_results} adds a partial-credit perspective. Many rollouts that fail the strict criterion still make substantial progress: for example, o3 leads state-repair score at 0.6757 despite only 33.7\% strict accuracy. This reveals that many failures are near-miss closures rather than early breakdowns. Models often identify the correct repair direction but miss one final side effect or replacement step, or stop as if the workflow were already complete after only a partial repair. The score view supports a useful failure taxonomy: early breakdown, correct direction but incomplete closure, stale-state mishandling, and hallucinated completion.

\subsubsection{Step Efficiency}

\Cref{tab:length_results} pairs reference-trajectory length with average executed steps, revealing two distinct failure styles. GPT-5.3-Codex and GPT-5.4 are step-efficient, staying near eight executed steps on the longest reference buckets while maintaining strong accuracy. Claude Opus 4.6 achieves higher strict accuracy on several buckets but uses substantially more steps (16--20 average steps on GT-6 through GT-8 tasks). This distinguishes two failure modes: \emph{step-efficient but imperfect} models reach compact, partially correct solutions but miss one final operation, while \emph{high-step low-conversion} models keep exploring without improving closure proportionally. In stateful workflows, extra steps are not neutral: they often correspond to repeated inspection, redundant list queries, or additional state mutations that do not improve the final evaluator-visible state. A benchmark that only reports pass/fail would collapse these two failure styles together; the combined accuracy/steps view shows that workflow competence involves not just whether a model eventually passes, but how efficiently it converts interaction budget into closure.

\subsubsection{Scenario-Level Discrimination}

\begin{figure*}[t]
    \centering
    \includegraphics[width=\linewidth]{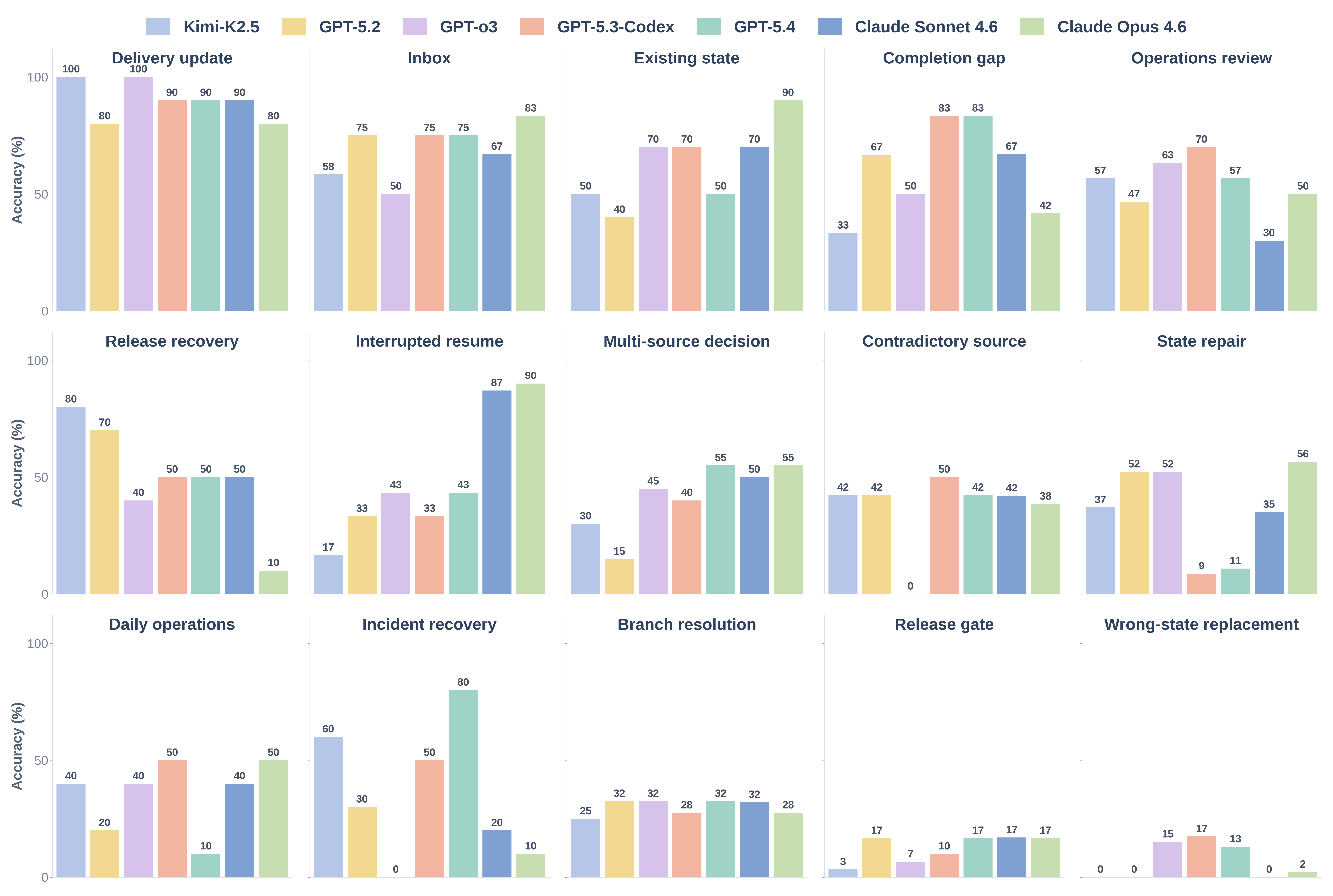}
    \caption{Per-scenario strict accuracy. Duplicate-aware scenarios are near saturation, while state repair, release gating, and wrong-state replacement remain challenging across all models.}
    \label{fig:scenario_breakdown}
    
\vspace{-15pt}
\end{figure*}

\Cref{fig:scenario_breakdown} shows that the benchmark's difficulty is not uniform. At one extreme, Already-Done Skip and Duplicate Avoidance are near saturation (90--100\% for most models) and are not shown in the figure. At the other extreme, Wrong-State Replacement remains below 17\% for all models, and Release Gate stays between 3\% and 17\%, exposing persistent weaknesses in state-conflict resolution. Between these extremes, Interrupted Workflow Resume produces the widest model separation (17\% to 90\%), with Claude models strongly outperforming GPT models on this slice. This per-scenario view demonstrates that the aggregate accuracy gap across models is not driven by uniform difficulty but by a small set of genuinely discriminative scenario families targeting state repair, replacement, and multi-step closure.

\subsection{Case Studies}
\label{sec:case_study}

The aggregate results above show \emph{where} models fail; this section illustrates \emph{how} they fail through two representative tasks. We select these cases because they test complementary state-conflict capabilities: one requires identifying and replacing incorrect state, while the other requires recognizing and preserving correct state. Together they cover the two hardest judgment calls in stateful workflows.

\noindent \textbf{Case 1: Wrong-State Replacement.}

\begin{tcolorbox}[colback=red!6,colframe=red!50!black,sharp corners,left=1mm,right=1mm,top=1mm,bottom=1mm]
\small
\textbf{Task.} \texttt{hard\_decision\_workflow\_261}: ``Something in the Seattle release setup was staged wrong. Check whether the next step is stale, replace the stale next step with the correct version, and leave the next step and sync correctly set.'' \\[4pt]
\textbf{Initial state.} The environment contains a pre-existing task \texttt{Existing Seattle release follow-up} (low priority, stale). The replacement task and replacement sync are both missing. \\[4pt]
\textbf{Required actions.} Inspect the seeded task, determine it is stale, retire it, create the corrected \texttt{Seattle release replacement next step}, and add the matching \texttt{Seattle replacement release sync} calendar event.\\[4pt]
\textbf{Evaluation.} Strict evaluation requires \emph{both} retirement of the stale task \emph{and} creation of the replacement state. Adding the replacement without cleanup, or cleaning up without recreating the follow-through, both fail.
\end{tcolorbox}

This task tests whether agents can identify and remove incorrect state before creating its replacement. In \Cref{tab:ability_results}, Wrong-State Replacement remains below 17\% strict accuracy for all models, making it the hardest shared slice in the benchmark. The partial-credit view (\Cref{tab:ability_score_results}) reveals that many models identify the stale object and attempt a replacement, but fail to complete the full retirement-and-recreation cycle, leaving the workflow in an inconsistent intermediate state. The common failure pattern is hallucinated completion: the model retires the stale task but stops before creating the replacement, or creates the replacement but leaves the stale task in place.

The second case tests the opposite judgment: rather than replacing wrong state, the agent must preserve correct state and only fill in what is missing.

\noindent \textbf{Case 2: Interrupted Workflow Resume.}

\begin{tcolorbox}[colback=cyan!6,colframe=cyan!50!black,sharp corners,left=1mm,right=1mm,top=1mm,bottom=1mm]
\small
\textbf{Task.} \texttt{hard\_decision\_workflow\_307}: ``The New York release work already has pieces in place. Check the current model, the next step on the board, and the sync on the calendar, finish the missing handoff file, send manager@example.com a recap, and leave the existing setup alone.'' \\[4pt]
\textbf{Initial state.} The environment already contains \texttt{New York existing release next step}, \texttt{New York existing release sync}, and drafted notes in \texttt{/ops/release-review.txt}. Only the handoff file and recap email are missing. \\[4pt]
\textbf{Required actions.} Resume from the partially completed state: read existing review notes, create the missing handoff file, preserve the staged task and sync (no duplicates), and send the recap email. \\[4pt]
\textbf{Evaluation.} The evaluator accepts only resume-style closure: the existing task and sync must remain, no duplicates may be created, the handoff file must appear, and the recap email must be sent.
\end{tcolorbox}

This task produces the widest model separation in \Cref{fig:scenario_breakdown}: Claude Opus 4.6 reaches 90\% while Kimi-K2.5 stays at 17\%. The gap is driven by whether the model inspects existing state before acting or immediately begins creating objects from scratch. Models that skip the inspection step tend to recreate the task and calendar event, triggering the duplicate guard and failing despite producing otherwise correct artifacts.

These two cases illustrate the central design principle of the benchmark. Static or clean-state benchmarks would score both tasks as straightforward multi-step completions. The difficulty arises entirely from pre-existing state: in Case~1 the agent must judge what is \emph{wrong} and replace it, while in Case~2 it must judge what is \emph{right} and leave it alone. This state-aware judgment is what the ClawForge-Bench is designed to measure. The full scenario catalog with additional examples is provided in Appendix~\ref{sec:appendix_catalog}.

\section{Related Work}
\label{sec:related}

\noindent \textbf{Interactive agent benchmarks.}
Recent work increasingly evaluates LLM agents as sequential decision-makers that interleave reasoning, tool use, and action rather than produce a single final answer, with ReAct serving as a representative formulation~\citep{yao2022react}. This shift has motivated executable or semi-executable benchmarks for software repair, web interaction, planning, and tool use, including SWE-bench~\citep{jimenez2023swe}, AgentBench~\citep{liu2023agentbench}, WebArena~\citep{zhou2023webarena}, OSWorld~\citep{xie2024osworld}, GAIA~\citep{mialon2023gaia}, Claw-Eval~\citep{ye2026claw}, and ClawsBench~\citep{li2026clawsbench}. However, these benchmarks mainly evaluate task completion or trajectory correctness under clean or weakly constrained initial state, making it difficult to measure failures caused by stale state, conflicting artifacts, or incomplete workflows.

\noindent \textbf{Persistent memory and long-horizon agents.}
A related line of work studies persistent memory and long-horizon behavior in LLM agents. Virtual-context approaches, including MEMGPT~\citep{packer2023memgpt}, MEMORYOS~\citep{kang2025memory}, and stream-based controllers~\citep{wang2024agent}, extend interaction length through memory paging or context management. Structured and graph-based systems, such as MEMORYBANK~\citep{zhong2024memorybank}, MEM0~\citep{chhikara2025mem0}, ZEP~\citep{rasmussen2025zep}, A-MEM~\citep{xu2025mem}, and O-MEM~\citep{wang2025mem}, instead organize memory into persistent representations for long-horizon retrieval and planning. These approaches primarily improve memory mechanisms, but provide limited evaluation of whether agents correctly preserve, repair, or update workflow state during extended execution.

\noindent \textbf{Stateful workflow evaluation.}
ClawForge focuses on evaluating these workflow-state behaviors directly. Unlike existing benchmarks that mainly evaluate final task completion or exact trajectories~\citep{li2026clawenvkit,ye2026claw}, ClawForge evaluates agents under persistent workflow state, where tasks may already be partially completed, inconsistent, stale, or duplicate-sensitive before execution begins. Agents must preserve valid state, detect outdated artifacts, resolve conflicting information, avoid redundant mutations, and complete workflows. Correctness is defined over normalized end states and observable side effects rather than exact action sequences, allowing multiple valid trajectories.
\section{Conclusion}

We presented ClawForge, a generator-backed benchmark framework that systematically evaluates command-line agents under state-conflict workflows through automated task construction, stateful execution, and result-first evaluation. The ClawForge-Bench (17 scenarios, 6 ability categories) exposes failure modes that static or clean-state evaluation cannot observe: wrong-state replacement remains below 17\% strict accuracy for all seven frontier models evaluated, and the widest model separation (17\% to 90\% on Interrupted Workflow Resume) is driven by whether agents inspect existing state before acting. Partial-credit and step-efficiency analyses further reveal that models exhibit qualitatively different failure styles, from step-efficient near-miss closures to high-step low-conversion exploration. Expanding scenario coverage to new workflow domains and reducing the manual effort for defining scenario templates are promising directions for future work.

\clearpage

{\small
\bibliography{main}
\bibliographystyle{icml2026}
}

\clearpage

\appendix
\onecolumn

\section{Benchmark Construction}
\label{sec:appendix_construction}

This section documents the scenario taxonomy, generated task objects, and implementation details that support reproducibility. We first present the full scenario inventory (\S\ref{sec:appendix_inventory}), then show a representative task instruction card (\S\ref{sec:appendix_instruction_card}), and finally describe the implementation contract (\S\ref{sec:appendix_implementation_detail}).

\subsection{Scenario Inventory}
\label{sec:appendix_inventory}

Table~\ref{tab:appendix_inventory} lists all 17 scenario families in the ClawForge-Bench, together with their primary ability classification, overlapping ability tags, and a brief description of the required workflow.

\begin{table*}
    \centering
    \footnotesize
    \caption{Full scenario inventory for \texttt{hard\_decision\_workflow}. Scenario names are rendered in paper form rather than as generator slugs; release counts are omitted because the generator allows per-scenario counts to be reconfigured.}
    \label{tab:appendix_inventory}
    \resizebox{\linewidth}{!}{%
    \begin{tabular}{p{0.24\textwidth} p{0.15\textwidth} p{0.24\textwidth} p{0.29\textwidth}}
        \toprule
        Scenario & Primary Ability & Ability Tags & Description \\
        \midrule
        Inbox & Information transfer & information transfer, workflow completion & Close an inbox thread, inspect board and calendar state, and add only the missing follow-up. \\
        Release Recovery Runbook & Workflow completion & workflow completion, schedule inference & Restore a partially staged release review by switching models and filling the missing review artifact. \\
        Channel Incident Recovery & Information transfer & information transfer, workflow completion & Send a channel update, verify tracking on the board, and add only the missing incident next step. \\
        Daily Operations Commitment Loop & Workflow completion & workflow completion, schedule inference & Use today's schedule plus board and cron state to fill the single missing daily operations commitment. \\
        Release Gate & Workflow completion & workflow completion, state inspection & Inspect model, board, and calendar state before closing the release gate cleanly. \\
        Delivery Update & Information transfer & information transfer, workflow completion & Repair a delivery path, avoid duplicate board work, and choose the right live or async closure. \\
        Operations Review & Workflow completion & workflow completion, schedule inference & Use forecast and schedule context to finish the missing operations setup before placing the review. \\
        Existing State & Gap completion & gap completion, state inspection, duplicate avoidance, workflow completion & Audit a partially completed recurring setup and fill exactly one missing piece without rebuilding the rest. \\
        Duplicate Avoidance & Duplicate avoidance & duplicate avoidance, completion recognition & Preserve existing operations setup and choose the correct primary or backup review block. \\
        Multi-Source Decision & Multi-source reasoning & multi-source reasoning, branch resolution, workflow completion & Combine email, forecast, and calendar evidence to decide between live and async closure. \\
        State Repair & State repair & state repair, state inspection, workflow completion & Detect one stale release artifact and repair it while preserving the rest of the setup. \\
        Completion Gap & Gap completion & gap completion, state inspection, workflow completion & Inspect a partial release setup and add only the missing task and/or sync. \\
        Branch Resolution & Multi-source reasoning & multi-source reasoning, branch resolution, duplicate avoidance & Resolve a live-vs-async branch from forecast and calendar context without duplicating setup. \\
        Already Done Skip & Duplicate avoidance & duplicate avoidance, completion recognition, information transfer & Verify that a release setup is already correct and send a recap without rebuilding anything. \\
        Wrong-State Replacement & State repair & state repair, replacement planning, workflow completion & Replace one wrongly staged artifact with the correct release state and preserve the valid pieces. \\
        Interrupted Workflow Resume & Gap completion & gap completion, state inspection, duplicate avoidance, workflow completion & Resume partially completed release work, finish the missing artifact, and avoid recreating existing state. \\
        Contradictory Source Resolution & Multi-source reasoning & multi-source reasoning, branch resolution, workflow completion, state inspection & Reconcile conflicting signals across sources before choosing the correct closure path. \\
        \bottomrule
    \end{tabular}%
    }
\end{table*}

\subsection{Example Task Instruction Card}
\label{sec:appendix_instruction_card}

To make the generated task format concrete, we show one complete instruction card from a gap-completion scenario. This illustrates how the five task components (instruction, initial state, required actions, checks, and difficulty) are coupled in a single specification.

\begin{tcolorbox}[myquote]
\textbf{Task ID.} \texttt{hard\_decision\_workflow\_204} (\texttt{completion\_gap\_followthrough}) \\[2pt]
\textbf{Primary ability.} Gap completion \\[2pt]
\textbf{Ability tags.} \texttt{gap\_completion}, \texttt{state\_inspection}, \texttt{workflow\_completion} \\[6pt]
\textbf{Instruction.} ``I already have part of the Berlin release work in motion, and the review sync is the missing piece. Review what's there, read the release notes, refresh the handoff file, leave the next step on the board in place and add only the missing sync on the calendar, and leave the rest alone.'' \\[6pt]
\textbf{Initial visible state.} The initial state already contains a pending board item for the Berlin release next step and a drafted release-review note. The calendar sync is still missing, and the handoff artifact needs to be created or refreshed. This means the task does not begin from an empty workspace: the agent must preserve valid state while identifying the one missing closure path. \\[6pt]
\textbf{Required correction and closure.} A semantically correct completion must keep the existing Berlin next-step task in place, create or refresh a handoff artifact, and add the missing Berlin release sync on the calendar. It must not create a duplicate next-step task. The reference trajectory uses the benchmark date anchor for Europe/Berlin, but the evaluator accepts alternative command traces as long as the final state closes the workflow without introducing duplicate board work. \\[6pt]
\textbf{Representative required checks.} This task mixes state-preservation and side-effect checks: the existing Berlin board item must still be present; a handoff-style file creation or refresh effect must be visible; no duplicate next-step task may be created; and the missing calendar sync must exist by the end of the rollout. \\[6pt]
\textbf{Why this task is hard.} The task is easy to understand semantically but easy to fail procedurally. Models can recognize that a review sync is missing and still fail by recreating the board item, omitting the handoff refresh, or stopping after scheduling the sync and incorrectly assuming the workflow is complete.
\end{tcolorbox}

\subsection{Implementation Detail}
\label{sec:appendix_implementation_detail}

This subsection describes how the task specification above is realized at runtime. At task level, each \texttt{hard\_decision\_workflow} episode is an executable specification rather than a bare prompt. The specification bundles five pieces of information: a natural-language instruction, scenario metadata, task-specific \texttt{initial\_state\_overrides}, a reference command trajectory, and typed evaluation checks. In practice, the task object also stores instruction variants, prompt style, realism and scenario tags, public or hidden task notes, and provider-related metadata used for analysis. At reset time, the environment loads a base state, applies the overrides declared by the task, initializes the backend for that state directory, and then exposes the instruction plus compact execution hints to the agent. For \texttt{hard\_decision\_workflow\_204}, the override layer inserts an already-existing Berlin next-step task and a drafted release-review note, so the evaluator can later distinguish correct gap completion from unnecessary re-creation.

During rollout, the environment records each command, its stdout/stderr, exit code, and inferred effects such as created files, completed tasks, or created calendar events. The rollout record therefore preserves not only what the agent typed, but what externally visible state transitions were actually produced. Scoring is then performed over a normalized result state rather than over exact trajectory imitation. This design matters for hard workflows because many tasks require a sequence of decisions before any mutation is safe: the agent often has to determine whether the relevant object already exists, whether it is stale, whether it should be preserved, and only then whether the correct action is to add, update, or replace state. In other words, the benchmark object is not just ``what command should be typed next,'' but ``what workflow state should exist when the rollout is over.''

The default benchmark runtime is \texttt{multi} mode, which uses explicit routing rather than one monolithic executor. Command families such as \texttt{openclaw}, \texttt{calendar}, \texttt{email}, \texttt{tasks}, \texttt{weather}, \texttt{file}, and \texttt{curl} are dispatched to their corresponding skills or adapters. This routing layer is important for reproducibility because it preserves cross-surface interaction while keeping the execution contract stable across benchmark runs.

One useful way to think about the implementation is as a four-stage contract: (1) construct initialized workflow state, (2) execute one command at a time under routed backends, (3) infer explicit state-changing effects from those executions, and (4) merge the resulting traces into evaluator-facing state. The appendix carries these details so that the main paper can focus on what the benchmark reveals about model behavior rather than on runtime mechanics alone.

\section{Evaluation Details}
\label{sec:appendix_evaluation}

This section shows how result-first scoring works in practice. We first describe the execution modes and evaluator internals (\S\ref{sec:appendix_exec_details}), then present representative evaluation checks (\S\ref{sec:appendix_check_examples}), and finally explain how to read the full scenario examples (\S\ref{sec:appendix_execution_examples}).

\subsection{Execution Mode and Evaluator Details}
\label{sec:appendix_exec_details}

Algorithm~\ref{alg:rollout} gives the episode-level rollout contract used by the main experiments.

\begin{algorithm}[H]
\caption{ClawForge episode rollout (detailed)}
\label{alg:rollout}
\begin{algorithmic}[1]
\STATE Load task $T$, base state $S_0$, and backend $B$
\STATE Apply task-specific state overrides and initialize evaluator $E$
\STATE Initialize observation $o_0 \leftarrow$ instruction, config, gateway status, command hints
\FOR{$t = 1, \dots, H$}
    \STATE Agent emits one command $a_t$
    \STATE Execute $a_t$ through backend $B$
    \STATE Update backend state and append inferred effects
    \STATE Construct next observation $o_t$
    \IF{$a_t \in \{\texttt{done}, \texttt{exit}, \texttt{quit}\}$ or $t = H$}
        \STATE break
    \ENDIF
\ENDFOR
\STATE Build normalized evaluation state $\hat{S}$
\STATE Return final result $E(\hat{S}, T)$
\end{algorithmic}
\end{algorithm}

ClawForge supports four execution modes with different fidelity and stability trade-offs:
\begin{itemize}
    \item \textbf{\texttt{mock}} uses a single in-process mock backend for unit tests and deterministic debugging.
    \item \textbf{\texttt{multi}} is the default benchmark setting, with explicit command-family routing through registered skills and adapters.
    \item \textbf{\texttt{real}} replaces the \texttt{openclaw} branch with a subprocess-backed real backend while keeping the same routing architecture.
    \item \textbf{\texttt{hybrid}} extends \texttt{real} with gateway lifecycle management, including auto-start, port injection, and health polling.
\end{itemize}

Internally, the evaluator builds a normalized state $\hat{S}$ that merges configuration state, gateway state, bounded command history, the latest stdout/stderr/exit code, and explicit effect traces (created calendar events, sent emails, created files, completed tasks, etc.). This normalization is particularly important for the hard benchmark because many tasks are judged by the relationship between pre-existing and newly created state: the evaluator must determine not only that an artifact exists, but whether an existing object was preserved, a stale object was retired, and the required downstream side effects were produced.

\subsection{Evaluation Check Examples}
\label{sec:appendix_check_examples}

Table~\ref{tab:appendix_checks} shows representative result-first checks from two hard tasks: \texttt{hard\_decision\_workflow\_204} (gap completion) and \texttt{hard\_decision\_workflow\_263} (wrong-state replacement). The first task rewards preserving correct existing state while filling a single missing gap. The second additionally requires retiring stale state and putting the replacement plan in place. This is why replacement-style tasks are materially harder: they require preserving the valid part of the workflow, modifying the invalid part, and still achieving end-to-end closure.

\begin{table*}
    \centering
    \footnotesize
    \caption{Representative task-level evaluation checks. These checks illustrate the benchmark's result-first philosophy: trajectories may differ, but the rollout must leave the required state and effects behind.}
    \label{tab:appendix_checks}
    \begin{tabular}{p{0.18\linewidth} p{0.38\linewidth} p{0.34\linewidth}}
        \toprule
        Check family & Concrete condition & Why it matters \\
        \midrule
        State (\texttt{204}) & Existing Berlin follow-through task remains present & Verifies that the rollout preserves already-correct board state rather than deleting or replacing it unnecessarily. \\
        Effect (\texttt{204}) & \texttt{files\_created count\_gte 1} and created path contains \texttt{handoff} & Accepts semantically correct handoff refreshes without forcing one exact filename. \\
        Effect (\texttt{204}) & \texttt{tasks\_created not\_exists} & Implements the no-duplicate guard for the existing release next step. \\
        Effect (\texttt{204}) & \texttt{calendar\_events\_created count\_gte 1} & Checks that the missing review sync was actually scheduled rather than merely mentioned. \\
        Effect (\texttt{263}) & \texttt{tasks\_completed count\_gte 1} & Requires the stale Paris follow-up to be retired rather than simply ignored. \\
        Effect (\texttt{263}) & Replacement task created and replacement sync created & Enforces replacement-and-closure: the corrected next step and the replacement sync must both exist. \\
        Output (both) & Final command exits successfully & Prevents a rollout from being counted as complete when the last required mutation failed. \\
        \bottomrule
    \end{tabular}
\end{table*}

Different checks can also carry different weights in the partial-credit score. In practice, this means the benchmark can distinguish between a rollout that gets the main task direction right but misses one final side effect, and a rollout that never reaches the correct repair or branch decision in the first place. The score layer is therefore not a soft alternative to evaluation, but an additional view on how close a failed rollout came to the intended state transition.

\subsection{Execution Examples}
\label{sec:appendix_execution_examples}

The appendix now separates two example roles. The later full scenario catalog gives one long-form example for every released scenario family, while the present section explains how to read those examples. Each long example follows the same schema: a concrete instance, the initial visible state, the key repair or branch decision, a short representative action trace, and the evaluator-facing success condition. This keeps the examples comparable even when one scenario is mostly about duplicate avoidance and another is mostly about state replacement or multi-source resolution.

A useful reading strategy is to compare three things across the catalog: first, whether the task begins from empty or partially populated state; second, whether the required operation is additive, repair-oriented, or replacement-oriented; and third, whether final closure depends on one surface or on coordination across board, calendar, email, file, and cron state. That contrast is exactly what makes the ClawForge-Bench analytically useful.

\section{Supplementary Results}
\label{sec:appendix_results}

This section provides additional experimental breakdowns that extend the main-text analysis: per-scenario accuracy slices (\S\ref{sec:appendix_scenario_slice}), recurring failure patterns (\S\ref{sec:appendix_failure_notes}), and step-efficiency interpretation (\S\ref{sec:appendix_efficiency_notes}).

\subsection{Scenario Slice Table}
\label{sec:appendix_scenario_slice}

Table~\ref{tab:appendix_results} highlights six scenario families that are particularly useful for interpreting the benchmark. The selected slices include near-saturated checks (\texttt{already\_done\_skip\_followthrough}, \texttt{duplicate\_avoidance\_followthrough}), a uniformly hard shared slice (\texttt{release\_gate\_followthrough}), the hardest replacement slice (\texttt{wrong\_state\_replacement\_followthrough}), a strong separator (\texttt{interrupted\_workflow\_resume}), and a multi-source branch-selection slice (\texttt{multi\_source\_decision\_followthrough}).

\begin{table*}
    \centering
    \caption{Supplementary scenario-level full-pass accuracy (\%) for selected hard slices. Values follow the same evaluated run set used in the main experiment section.}
    \label{tab:appendix_results}
    \footnotesize
    \setlength{\tabcolsep}{4pt}
    \resizebox{\linewidth}{!}{%
    \begin{tabular}{lccccccc}
        \toprule
        Scenario & Kimi-K2.5 & GPT-5.2 & o3 & GPT-5.3-Codex & GPT-5.4 & Claude Sonnet 4.6 & Claude Opus 4.6 \\
        \midrule
        already\_done\_skip\_followthrough & 100 & 100 & 90 & 90 & 100 & 100 & 100 \\
        duplicate\_avoidance\_followthrough & 90 & 90 & 80 & 100 & 100 & 100 & 90 \\
        release\_gate\_followthrough & 3 & 17 & 7 & 10 & 17 & 17 & 17 \\
        wrong\_state\_replacement\_followthrough & 0 & 0 & 7 & 17 & 13 & 0 & 2 \\
        interrupted\_workflow\_resume & 17 & 33 & 47 & 33 & 43 & 87 & 90 \\
        multi\_source\_decision\_followthrough & 30 & 15 & 55 & 40 & 55 & 50 & 55 \\
        \bottomrule
    \end{tabular}%
    }
\end{table*}

\subsection{Failure Pattern Notes}
\label{sec:appendix_failure_notes}

Beyond per-scenario accuracy, the benchmark reveals recurring failure patterns that cut across scenario families. The main text argues that many failures are structured rather than arbitrary; Table~\ref{tab:appendix_failure_patterns} expands that claim with concrete pattern-level categories that recur across scenario families.

\begin{table*}
    \centering
    \footnotesize
    \caption{Representative failure patterns that recur across the hard benchmark.}
    \label{tab:appendix_failure_patterns}
    \begin{tabular}{p{0.22\linewidth} p{0.33\linewidth} p{0.35\linewidth}}
        \toprule
        Failure pattern & Typical benchmark symptom & Why it matters analytically \\
        \midrule
        Early breakdown & Wrong branch or wrong object selected before meaningful state change & Indicates misunderstanding of the task logic rather than near-miss closure. \\
        Incomplete closure & Main repair direction is correct, but one final side effect is absent & Distinguishes almost-finished rollouts from fully incorrect ones. \\
        Stale-state mishandling & Valid state is overwritten, or stale state is left in place while new state is added & Captures failures in state conflict resolution rather than in simple completion. \\
        Hallucinated completion & Agent stops after partial repair and behaves as if the workflow were done & Explains why some runs earn high score but still fail strict evaluation. \\
        Excess-step inefficiency & Rollout consumes many extra steps without proportionate progress toward final closure & Exposes weak workflow efficiency rather than just low pass rate. \\
        \bottomrule
    \end{tabular}
\end{table*}

\subsection{Length and Efficiency Notes}
\label{sec:appendix_efficiency_notes}

The previous subsections examine what models get wrong; this subsection examines how efficiently they use their interaction budget. The length-stratified analysis in the main paper pairs reference-trajectory length with average executed steps. A useful interpretation is that GT length measures the structural size of the intended workflow, while executed steps measure how much interaction budget a model actually spends on that workflow. These quantities need not match closely. A model can be step-efficient but still miss one final operation, or it can consume many additional steps without improving final closure.

This distinction is particularly important for stateful tasks. In many ClawForge-Bench episodes, extra steps are not neutral: they often correspond to repeated inspection, redundant list queries, or additional state mutations that do not improve the final evaluator-visible state. The benchmark therefore treats step usage as a behavioral signal about workflow efficiency, not just as an implementation detail of the rollout trace.

\section{Data and Reproducibility}
\label{sec:appendix_data}

\subsection{Data Provenance and Provider Impact}
\label{sec:appendix_data_provenance}

The main experimental tables report benchmark outcomes over the shared ClawForge-Bench snapshot. For Kimi-K2.5, GPT-5.2, GPT-o3, GPT-5.3-Codex, GPT-5.4, and Claude Opus 4.6, the reported strict accuracy, partial-credit score, and length-bucket average steps are computed from task-level traces over the evaluated run set. The Claude Models aggregate accuracy is reported over the same task denominator, but the length-bucket average-step values are computed from the available task-level traces selected for the merged available run. 

Provider-aware reporting is included because benchmark runs over external model endpoints can introduce retries, filtered outputs, or fallback behavior that affect an otherwise valid interactive rollout. In the reporting scheme, \texttt{provider\_failures} counts tasks that fail outright because a provider-side issue prevents a normal completion. \texttt{provider\_impacted\_tasks} counts tasks whose trajectories complete but are materially disturbed by provider-side effects such as compact fallbacks or filtered responses. These statistics are explanatory rather than primary: they do not replace full-pass accuracy or partial-credit score, but help readers judge whether serving noise is large enough to affect a comparison. The intended reading order is therefore to examine accuracy and score first, use the length table to interpret closure efficiency, and then use provider-aware statistics to determine whether a model's execution trace was unusually noisy at the serving layer.

\subsection{Limitations and Scope}
\label{sec:appendix_limitations}

Several limitations bound the claims made in the main paper. First, the benchmark is intentionally narrow in environment type: it studies executable command-line workflow episodes over tasks, calendar, email, messaging, files, configuration, weather, and cron state, rather than arbitrary long-horizon agent deployment. The intended claim is therefore not that one benchmark exhausts real agent reliability, but that state-conflict workflows expose failure modes that static prompting and lighter tool-use tests often miss.

Second, the reported model comparisons are benchmark evaluations over one shared ClawForge-Bench snapshot rather than variance estimates over many independently repeated runs. This matters especially for externally served agents, where provider-side retries, filtered outputs, and compact-fallback behavior can perturb a rollout even when the task specification is fixed. The paper surfaces those effects through provider-aware accounting, but it does not eliminate them.

Third, the hard suite emphasizes result-first closure under persistent state, so some conclusions are benchmark-specific by construction. The strongest separations in this paper come from duplicate-sensitive completion, stale-state repair, replacement, and branch resolution. Other capabilities that matter in agent deployment, such as open-ended research, web navigation, or multimodal perception, are outside the scope of this benchmark family.

Fourth, the Sonnet 4.6 reporting line is not perfectly symmetric with the other entries. Its strict-accuracy value uses the shared task denominator, but some diagnostic step statistics rely on the available merged best-available traces rather than one fully uniform trace set. The main text and \Cref{sec:appendix_data_provenance} therefore treat those step counts as efficiency evidence, not as a clean latency benchmark.

\subsection{Statistical Significance}
\label{sec:appendix_significance}

The paper reports deterministic benchmark outcomes on a fixed task snapshot and emphasizes exact full-pass accuracy, partial-credit score, and structured breakdown tables. It does not report confidence intervals, bootstrap uncertainty, or hypothesis tests. This omission is deliberate rather than hidden: the current paper is positioned as a benchmark-construction and empirical-separation study, not as a statistical comparison of nearly tied models under repeated random resampling.

The absence of formal significance reporting should shape how close comparisons are read. Large gaps tied to stable scenario families, such as the persistent difficulty of replacement-style tasks or the near-saturation of already-done and duplicate-avoidance slices, are more important to the paper's argument than tiny differences between adjacent aggregate rankings. Future versions of the benchmark can add repeated-run variance estimation or bootstrap intervals over tasks, but that analysis is not part of the present submission.

\subsection{Compute Resources}
\label{sec:appendix_compute}

The reported experiments do not train new models. The local benchmark workload is therefore orchestration-heavy rather than training-heavy: task reset, routed command execution, evaluator-state construction, and result aggregation all run through the ClawForge runtime in \texttt{multi} mode with full interaction history and a 25-step budget. In this setup, the benchmark runner itself is CPU-oriented and does not require local GPU execution for the paper's reported results.

\section{Broader Impact and Release}
\label{sec:appendix_impact}

\subsection{Broader Impact}
\label{sec:appendix_broader_impacts}

\textbf{Positive impact.}
This work improves measurement quality for interactive agents. Many real-world failures arise from state conflicts rather than one-shot misunderstanding. By explicitly capturing failure modes such as duplication, stale state, incorrect branching, and incomplete workflow closure, the benchmark enables more precise diagnosis of where agents remain unreliable.

\textbf{Risks.}
Improved benchmark coverage may indirectly accelerate deployment of agents in coordination, operations, or messaging workflows, where incorrect execution can incur real-world cost. Benchmark progress may also be misinterpreted as readiness for broader automation beyond the evaluated scenarios.

\textbf{Misuse and mitigation.}
Workflow benchmarks can be used to optimize agents that manipulate calendars, messages, files, or system state, including in settings with limited oversight. To mitigate this, the benchmark is built on synthetic or seeded environments, emphasizes transparent reporting of failure modes, and frames evaluation as a diagnostic tool rather than evidence of safe autonomous deployment.

\subsection{Asset Credits and Terms}
\label{sec:appendix_asset_terms}

The paper uses two main classes of existing assets: prior benchmark literature and externally served foundation-model endpoints. Prior interactive-benchmark and agent-evaluation papers are cited in the introduction, method, experiment, and related-work sections. The evaluated models are likewise credited by provider and citation: Kimi-K2.5~\citep{team2026kimi} to the Kimi team, GPT~\citep{openai2026gpt54} to OpenAI, and Claude Code~\cite{authropic3} models to Anthropic.

These evaluated models are not treated as open-weight assets in this paper. They are proprietary provider endpoints accessed under the corresponding provider's service terms rather than redistributed model weights. The paper therefore makes a sharper distinction than many open-model evaluations: for these systems, the relevant usage condition is provider-controlled API or hosted-access terms, not an open-source model license embedded in the benchmark package itself.

The new benchmark artifacts described in this paper are ClawForge task specifications, seeded state, evaluator contracts, and reporting views. They are documented as new assets of the study rather than borrowed benchmark files relabeled as original contributions. The paper does not claim to redistribute third-party model weights, proprietary datasets, or third-party application data snapshots.

\subsection{Open Access Release Status}
\label{sec:appendix_open_access}

This submission documents the benchmark protocol, task structure, evaluator contract, and reporting methodology in enough detail to support scientific interpretation, but it does not attach an anonymized open-access release bundle for the benchmark snapshot and code at submission time. That is a release-status fact, not a hidden omission.

The practical reason is that a faithful release package for this benchmark is more than a PDF appendix: it must include task specifications, state initialization assets, command-routing runtime code, evaluator definitions, and clear instructions for running the shared hard snapshot against supported model providers. The paper already documents those components conceptually, but the anonymized distribution bundle is not part of the present submission package.

\section{Full Scenario Catalog}
\label{sec:appendix_catalog}

\subsection{Scenario Example Directory}
\label{sec:appendix_example_directory}

\begin{table*}
    \centering
    \footnotesize
    \caption{Directory of the full scenario example catalog. Page references point to the start of each long example.}
    \label{tab:scenario_example_guide}
    \resizebox{\linewidth}{!}{%
    \begin{tabular}{p{0.34\linewidth} p{0.14\linewidth} p{0.34\linewidth} p{0.14\linewidth}}
        \toprule
        Scenario & Page & Scenario & Page \\
        \midrule
        Inbox & \pageref{app:ex-inbox} & Multi-Source Decision & \pageref{app:ex-multi-source-decision} \\
        Release Recovery Runbook & \pageref{app:ex-release-recovery-runbook} & State Repair & \pageref{app:ex-state-repair} \\
        Channel Incident Recovery & \pageref{app:ex-channel-incident-recovery} & Completion Gap & \pageref{app:ex-completion-gap} \\
        Daily Operations Commitment Loop & \pageref{app:ex-daily-operations-commitment-loop} & Branch Resolution & \pageref{app:ex-branch-resolution} \\
        Release Gate & \pageref{app:ex-release-gate} & Already Done Skip & \pageref{app:ex-already-done-skip} \\
        Delivery Update & \pageref{app:ex-delivery-update} & Wrong-State Replacement & \pageref{app:ex-wrong-state-replacement} \\
        Operations Review & \pageref{app:ex-operations-review} & Interrupted Workflow Resume & \pageref{app:ex-interrupted-workflow-resume} \\
        Existing State & \pageref{app:ex-existing-state} & Contradictory Source Resolution & \pageref{app:ex-contradictory-source-resolution} \\
        Duplicate Avoidance & \pageref{app:ex-duplicate-avoidance} &  &  \\
        \bottomrule
    \end{tabular}%
    }
\end{table*}

\subsection{Full Scenario Example Catalog}
\label{sec:appendix_full_catalog}

The following catalog now uses one concrete representative task instance for every released scenario family. Each box is grounded in one actual \texttt{hard\_decision\_workflow\_xxx} specification rather than in a scenario-level summary, so the instruction wording, seeded state, representative commands, and evaluator-visible closure conditions all come from a real released task.

\begin{tcolorbox}[colback=blue!3,colframe=blue!25,title={{\small Example A: Inbox (\texttt{hard\_decision\_workflow\_1})}},fonttitle=\bfseries\small,left=6pt,right=6pt,top=4pt,bottom=4pt]
\small\label{app:ex-inbox}
\textbf{Representative instance.} \texttt{hard\_decision\_workflow\_1} asks: ``The budget email thread in the inbox needs attention today. Then check the Berlin board and calendar; right now the next step is already on the board, and if it only needs async follow-up, leave the calendar alone, so fill only what is missing, note it in a handoff file, and send alice@example.com a quick update.'' \\[4pt]
\textbf{Initial state.} The seeded board already contains the pending task \texttt{Berlin budget follow-up}. The task instruction also implies that the follow-through should stay async if the existing calendar state already supports that branch, so the missing work is the handoff note plus the external update rather than a new calendar block. \\[4pt]
\textbf{Required correction.} Read the budget thread, verify that the staged Berlin task can stay in place, avoid adding a duplicate task or review event, create the handoff note under \texttt{/handoff}, and send the requested email update to \texttt{alice@example.com}. \\[4pt]
\textbf{Representative actions.}
\begin{enumerate}[leftmargin=18pt, itemsep=2pt, topsep=2pt]
    \item \texttt{email search --query 'budget'}
    \item \texttt{email read --id email\_seed\_3}
    \item \texttt{tasks list --status pending}
    \item \texttt{calendar list}
    \item \texttt{calendar today --timezone Europe/Berlin}
    \item \texttt{file create --path '/handoff/berlin-budget-followup.txt' --content 'Berlin budget follow-up handoff note.'}
    \item \texttt{email send --to alice@example.com --subject 'Budget follow-up for Berlin' --body 'Tracking the next step for Berlin.'}
\end{enumerate}
\tcblower
\small
\textbf{Evaluation outcome.} The evaluator accepts this rollout only if the handoff file is created, the update email is sent, the pre-existing Berlin follow-up task remains sufficient, and no duplicate task or calendar event is introduced. This makes the task a real inbox-grounded completion problem rather than a generic email summary.
\end{tcolorbox}

\begin{tcolorbox}[colback=green!3,colframe=green!30!black,title={{\small Example B: Release Recovery Runbook (\texttt{hard\_decision\_workflow\_7})}},fonttitle=\bfseries\small,left=6pt,right=6pt,top=4pt,bottom=4pt]
\small\label{app:ex-release-recovery-runbook}
\textbf{Representative instance.} \texttt{hard\_decision\_workflow\_7} asks: ``Please tighten up the Berlin release review. Check what's already on the board and calendar, use anthropic/claude-opus-4-6, refresh the handoff file, and add only the missing review slot or decision-log task.'' \\[4pt]
\textbf{Initial state.} The override layer already stages a pending Berlin release next-step task, so the workflow is not empty. What is missing is the corrected review configuration under the required model path together with a refreshed handoff artifact. \\[4pt]
\textbf{Required correction.} Confirm the current model, switch to \texttt{anthropic/claude-opus-4-6}, preserve the existing Berlin release next step, refresh the handoff file, and add only the missing Berlin review slot rather than recreating the release workflow from scratch. \\[4pt]
\textbf{Representative actions.}
\begin{enumerate}[leftmargin=18pt, itemsep=2pt, topsep=2pt]
    \item \texttt{openclaw config get agent.model}
    \item \texttt{tasks list --status pending}
    \item \texttt{calendar list}
    \item \texttt{calendar today --timezone Europe/Berlin}
    \item \texttt{openclaw models set anthropic/claude-opus-4-6}
    \item \texttt{file create --path '/ops/release-handoff.txt' --content 'Berlin release handoff notes.'}
    \item \texttt{calendar add-event --title 'Berlin release review' --start 2026-03-10T09:00}
\end{enumerate}
\tcblower
\small
\textbf{Evaluation outcome.} A passing rollout must visibly set the model to the required Opus path, create the handoff file, keep the existing decision-log reminder and board task intact, and place the Berlin review event on the calendar. The task fails if the model switch is skipped or if valid staged state is rebuilt unnecessarily.
\end{tcolorbox}

\begin{tcolorbox}[colback=orange!4,colframe=orange!45!black,title={{\small Example C: Channel Incident Recovery (\texttt{hard\_decision\_workflow\_13})}},fonttitle=\bfseries\small,left=6pt,right=6pt,top=4pt,bottom=4pt]
\small\label{app:ex-channel-incident-recovery}
\textbf{Representative instance.} \texttt{hard\_decision\_workflow\_13} asks: ``Please get the incident update to \#general on discord, check the incident next step on the board first to see whether it is already there, add only the missing piece, and send alice@example.com a recap when it's handled.'' \\[4pt]
\textbf{Initial state.} The board already contains the pending task \texttt{Incident escalation follow-up}. The missing work is therefore the externally visible incident post plus the recap email, not another board task. \\[4pt]
\textbf{Required correction.} Audit the task board first, post the incident escalation update to Discord \texttt{\#general}, avoid recreating the follow-up task that already exists, and send the requested recap to \texttt{alice@example.com}. \\[4pt]
\textbf{Representative actions.}
\begin{enumerate}[leftmargin=18pt, itemsep=2pt, topsep=2pt]
    \item \texttt{openclaw security audit}
    \item \texttt{tasks list --status pending}
    \item \texttt{openclaw channels login --channel discord}
    \item \texttt{openclaw message send --channel discord --target \#general --message 'Incident escalation started. Please acknowledge.'}
    \item \texttt{openclaw channels list --json}
    \item \texttt{email send --to alice@example.com --subject 'Incident escalation recap' --body 'The escalation is posted and being tracked.'}
\end{enumerate}
\tcblower
\small
\textbf{Evaluation outcome.} Strict evaluation requires both external side effects: the Discord message to \texttt{\#general} and the recap email. It also checks that the existing follow-up task remains the only board artifact. This makes the task a clean example of information transfer plus duplicate-aware closure.
\end{tcolorbox}

\begin{tcolorbox}[colback=cyan!3,colframe=cyan!35!black,title={{\small Example D: Daily Operations Commitment Loop (\texttt{hard\_decision\_workflow\_19})}},fonttitle=\bfseries\small,left=6pt,right=6pt,top=4pt,bottom=4pt]
\small\label{app:ex-daily-operations-commitment-loop}
\textbf{Representative instance.} \texttt{hard\_decision\_workflow\_19} asks: ``Can you check today's Europe/Berlin schedule for Berlin, look at the next step on the board and the recurring daily check, and only fill the missing piece?'' \\[4pt]
\textbf{Initial state.} The seeded state already contains \texttt{Berlin existing ops next-step task}. No recurring daily-check cron is present yet, so only the recurring commitment is missing. \\[4pt]
\textbf{Required correction.} Inspect the Berlin schedule, verify the existing next-step task, check the current cron state, and add only the missing recurring daily-check job without touching the valid board state. \\[4pt]
\textbf{Representative actions.}
\begin{enumerate}[leftmargin=18pt, itemsep=2pt, topsep=2pt]
    \item \texttt{weather forecast --location 'Berlin' --days 3}
    \item \texttt{calendar today --timezone Europe/Berlin}
    \item \texttt{tasks list --status pending}
    \item \texttt{openclaw cron list}
    \item \texttt{openclaw cron add --name hard-ops-01 --cron '0 9 * * *' --message 'Run Berlin daily ops check'}
\end{enumerate}
\tcblower
\small
\textbf{Evaluation outcome.} This rollout passes only if the Berlin next-step task remains in place and exactly one new cron job is created for the daily ops check. It fails when the model duplicates board work, adds the wrong recurring artifact, or stops after inspection without committing the missing scheduler state.
\end{tcolorbox}

\begin{tcolorbox}[colback=purple!3,colframe=purple!30!black,title={{\small Example E: Release Gate (\texttt{hard\_decision\_workflow\_43})}},fonttitle=\bfseries\small,left=6pt,right=6pt,top=4pt,bottom=4pt]
\small\label{app:ex-release-gate}
\textbf{Representative instance.} \texttt{hard\_decision\_workflow\_43} asks: ``Please get the Singapore release gate sorted out. Confirm the current model, review the board and calendar first, then update the handoff file, the next step on the board, and the sync on the calendar.'' \\[4pt]
\textbf{Initial state.} This instance does not preload the release artifacts through overrides, so the full gate closure must be assembled from inspection plus creation. The hidden constraints require a specific model path, a handoff file, a release-gate follow-through task, and a matching sync. \\[4pt]
\textbf{Required correction.} Confirm the current model, inspect board and calendar state, switch to \texttt{anthropic/claude-3-7-sonnet-latest}, create the release handoff file, create the Singapore release-gate task, and place the release-gate sync on the calendar. \\[4pt]
\textbf{Representative actions.}
\begin{enumerate}[leftmargin=18pt, itemsep=2pt, topsep=2pt]
    \item \texttt{openclaw config get agent.model}
    \item \texttt{tasks list --status pending}
    \item \texttt{calendar list}
    \item \texttt{calendar today --timezone Asia/Singapore}
    \item \texttt{openclaw models set anthropic/claude-3-7-sonnet-latest}
    \item \texttt{file create --path '/ops/release-handoff.txt' --content 'Singapore release gate handoff notes.'}
    \item \texttt{tasks add --title 'Singapore release gate follow-through' --priority high --due 2026-03-08}
    \item \texttt{calendar add-event --title 'Singapore release gate sync' --start 2026-03-10T09:00}
    \item \texttt{calendar today --timezone Asia/Singapore}
\end{enumerate}
\tcblower
\small
\textbf{Evaluation outcome.} Evaluation checks all four closure outputs together: model set, handoff file created, task created, and calendar event created. The task therefore exposes whether the agent can carry a release-gate workflow through to a fully consistent final state rather than stopping after one local fix.
\end{tcolorbox}

\begin{tcolorbox}[colback=yellow!4,colframe=yellow!50!black,title={{\small Example F: Delivery Update (\texttt{hard\_decision\_workflow\_73})}},fonttitle=\bfseries\small,left=6pt,right=6pt,top=4pt,bottom=4pt]
\small\label{app:ex-delivery-update}
\textbf{Representative instance.} \texttt{hard\_decision\_workflow\_73} asks: ``Can you get the outage update out to \#launch on discord? Fix the delivery path if needed, check what's already on the board, the next step may already be on the board, so leave it alone if it is there, if the target is shared, leave a short live block on the calendar, and send bob@example.com a quick recap.'' \\[4pt]
\textbf{Initial state.} The board already contains \texttt{Outage delivery next step}. The required shared-target branch also expects a short live calendar block, but neither the Discord post nor the recap email has been sent yet. \\[4pt]
\textbf{Required correction.} Leave the staged outage next step alone, post the outage update to Discord \texttt{\#launch}, add the live follow-up calendar block, and send the recap to \texttt{bob@example.com}. \\[4pt]
\textbf{Representative actions.}
\begin{enumerate}[leftmargin=18pt, itemsep=2pt, topsep=2pt]
    \item \texttt{tasks list --status pending}
    \item \texttt{openclaw channels list --json}
    \item \texttt{openclaw channels login --channel discord}
    \item \texttt{calendar add-event --title 'Outage delivery follow-up' --start 2026-03-12T14:00}
    \item \texttt{openclaw message send --channel discord --target \#launch --message 'Outage update posted and being tracked.'}
    \item \texttt{email send --to bob@example.com --subject 'Outage delivery recap' --body 'The delivery path is recovered and the update is posted.'}
\end{enumerate}
\tcblower
\small
\textbf{Evaluation outcome.} The evaluator requires all three visible side effects while also enforcing that no duplicate board task is created. This task is representative because it mixes messaging, scheduling, and duplicate-aware state preservation inside one delivery workflow.
\end{tcolorbox}

\begin{tcolorbox}[colback=teal!3,colframe=teal!35!black,title={{\small Example G: Operations Review (\texttt{hard\_decision\_workflow\_83})}},fonttitle=\bfseries\small,left=6pt,right=6pt,top=4pt,bottom=4pt]
\small\label{app:ex-operations-review}
\textbf{Representative instance.} \texttt{hard\_decision\_workflow\_83} asks: ``Can you look at the forecast and today's Europe/London calendar for London, review the next step on the board and the recurring daily check, and only add what's missing before you put the review on the calendar?'' \\[4pt]
\textbf{Initial state.} The seeded state already contains \texttt{London existing ops next-step task}. The missing pieces are the recurring daily-check cron and the review event itself, both conditioned on current forecast and London calendar context. \\[4pt]
\textbf{Required correction.} Read the forecast, inspect the London calendar, confirm the existing ops task, create the missing daily-check cron, and place the \texttt{London backup ops review} event on the calendar. \\[4pt]
\textbf{Representative actions.}
\begin{enumerate}[leftmargin=18pt, itemsep=2pt, topsep=2pt]
    \item \texttt{weather forecast --location 'London' --days 1}
    \item \texttt{calendar today --timezone Europe/London}
    \item \texttt{openclaw cron list}
    \item \texttt{tasks list --status pending}
    \item \texttt{openclaw cron add --name hard-ops-07-01 --cron '15 8 * * 1-5' --message 'Run London daily ops check'}
    \item \texttt{calendar add-event --title 'London backup ops review' --start 2026-03-10T13:00}
\end{enumerate}
\tcblower
\small
\textbf{Evaluation outcome.} A passing rollout must keep the existing task intact while creating both the cron job and the review event. The task fails if the model schedules a review without completing the missing recurring setup or if it adds duplicate board state while trying to be safe.
\end{tcolorbox}

\begin{tcolorbox}[colback=magenta!3,colframe=magenta!35!black,title={{\small Example H: Existing State (\texttt{hard\_decision\_workflow\_113})}},fonttitle=\bfseries\small,left=6pt,right=6pt,top=4pt,bottom=4pt]
\small\label{app:ex-existing-state}
\textbf{Representative instance.} \texttt{hard\_decision\_workflow\_113} asks: ``The Boston daily ops setup is only partway done. Look over the next step on the board, the review slot on the calendar, and the recurring daily check, then finish just the missing piece and send alice@example.com a short recap.'' \\[4pt]
\textbf{Initial state.} The override layer preloads both \texttt{Boston existing ops next-step task} and \texttt{Boston existing ops review block}. The only missing piece is the recurring daily-check cron, after which a recap email must be sent. \\[4pt]
\textbf{Required correction.} Audit the partially completed Boston ops setup, preserve the existing task and review block, create only the missing cron job, and send the recap to \texttt{alice@example.com}. \\[4pt]
\textbf{Representative actions.}
\begin{enumerate}[leftmargin=18pt, itemsep=2pt, topsep=2pt]
    \item \texttt{openclaw cron list}
    \item \texttt{tasks list --status pending}
    \item \texttt{calendar list}
    \item \texttt{calendar today --timezone America/New\_York}
    \item \texttt{openclaw cron add --name existing-hard-ops-01 --cron '30 9 * * 1-5' --message 'Run Boston daily ops check'}
    \item Send the Boston ops recap email to \texttt{alice@example.com}.
\end{enumerate}
\tcblower
\small
\textbf{Evaluation outcome.} Strict evaluation checks that the existing task and review block remain present, no duplicate task or calendar event appears, the cron job is created, and the recap email is sent. This makes the example a concrete existing-state completion task rather than a vague recurring-workflow summary.
\end{tcolorbox}

\begin{tcolorbox}[colback=lime!3,colframe=lime!35!black,title={{\small Example I: Duplicate Avoidance (\texttt{hard\_decision\_workflow\_123})}},fonttitle=\bfseries\small,left=6pt,right=6pt,top=4pt,bottom=4pt]
\small\label{app:ex-duplicate-avoidance}
\textbf{Representative instance.} \texttt{hard\_decision\_workflow\_123} asks: ``Some of the London ops setup is already in place. Check the board, forecast, and calendar, then add a backup review block on the calendar if the forecast looks risky; otherwise add a primary review block on the calendar, and avoid recreating the existing setup.'' \\[4pt]
\textbf{Initial state.} The workflow already contains \texttt{London existing ops next-step task} and an active recurring cron \texttt{london-existing-daily-ops-check}. The only missing artifact is the review block, whose title depends on the forecast branch. \\[4pt]
\textbf{Required correction.} Inspect the existing board and cron state, read the London forecast and calendar, choose the backup review branch supported by the task instance, and add only the \texttt{London backup ops review block} without recreating existing setup. \\[4pt]
\textbf{Representative actions.}
\begin{enumerate}[leftmargin=18pt, itemsep=2pt, topsep=2pt]
    \item \texttt{openclaw cron list}
    \item \texttt{tasks list --status pending}
    \item \texttt{calendar list}
    \item \texttt{weather forecast --location 'London' --days 1}
    \item \texttt{calendar today --timezone Europe/London}
    \item \texttt{calendar add-event --title 'London backup ops review block' --start 2026-03-20T08:30}
\end{enumerate}
\tcblower
\small
\textbf{Evaluation outcome.} The evaluator rewards recognition that most of the workflow already exists. It checks that the board task and daily check stay in place, no duplicate task or cron job is added, and the review event is created with a backup-oriented title.
\end{tcolorbox}

\begin{tcolorbox}[colback=gray!7,colframe=gray!45,title={{\small Example J: Multi-Source Decision (\texttt{hard\_decision\_workflow\_133})}},fonttitle=\bfseries\small,left=6pt,right=6pt,top=4pt,bottom=4pt]
\small\label{app:ex-multi-source-decision}
\textbf{Representative instance.} \texttt{hard\_decision\_workflow\_133} asks: ``For Austin, check the review email note, the forecast, and the America/Chicago calendar, decide whether this stays live or shifts async, leave the next step on the board, keep the daily check scheduled, put the review on the calendar if live, and send leadership@example.com an update only if async.'' \\[4pt]
\textbf{Initial state.} This Austin instance begins without task-specific overrides. The closure branch must be inferred from the review email note, the forecast, and the America/Chicago calendar, and the live branch requires creating the task, cron job, and calendar review event. \\[4pt]
\textbf{Required correction.} Read the review email note, inspect the forecast and Chicago calendar, decide that the workflow stays live, create the Austin next-step task, keep the daily check scheduled, and place the \texttt{Austin primary ops review} event on the calendar. \\[4pt]
\textbf{Representative actions.}
\begin{enumerate}[leftmargin=18pt, itemsep=2pt, topsep=2pt]
    \item \texttt{weather forecast --location 'Austin' --days 1}
    \item \texttt{calendar today --timezone America/Chicago}
    \item \texttt{email search --query 'review'}
    \item \texttt{email read --id email\_seed\_5}
    \item \texttt{openclaw cron list}
    \item \texttt{tasks list --status pending}
    \item \texttt{tasks add --title 'Austin ops next step' --priority medium --due 2026-03-08}
    \item \texttt{openclaw cron add --name multi-source-hard-ops-01 --cron '0 9 * * *' --message 'Run Austin daily ops check'}
    \item \texttt{calendar add-event --title 'Austin primary ops review' --start 2026-03-10T09:00}
\end{enumerate}
\tcblower
\small
\textbf{Evaluation outcome.} The evaluator requires task creation, cron creation, and review-event creation together. A rollout that sends an async email instead of committing to the live branch, or that keeps inspecting without deciding, still fails even if the intermediate reasoning looks plausible.
\end{tcolorbox}

\begin{tcolorbox}[colback=red!4,colframe=red!45!black,title={{\small Example K: State Repair (\texttt{hard\_decision\_workflow\_153})}},fonttitle=\bfseries\small,left=6pt,right=6pt,top=4pt,bottom=4pt]
\small\label{app:ex-state-repair}
\textbf{Representative instance.} \texttt{hard\_decision\_workflow\_153} asks: ``The Boston release setup has something stale in it. Review the next-step task on the board and the sync on the calendar, repair the stale next step, refresh the handoff file, and leave the next step and sync correctly set.'' \\[4pt]
\textbf{Initial state.} The seeded state contains a stale low-priority task, \texttt{Existing Boston release follow-up}, under the \texttt{hard\_wrong\_task} setup. The correct release follow-through, handoff refresh, and Boston release sync do not yet exist. \\[4pt]
\textbf{Required correction.} Detect that the seeded Boston follow-up is stale, refresh the handoff file, complete the outdated task, create the corrected high-priority task \texttt{Boston release follow-through}, and add the \texttt{Boston release sync} calendar event. \\[4pt]
\textbf{Representative actions.}
\begin{enumerate}[leftmargin=18pt, itemsep=2pt, topsep=2pt]
    \item \texttt{tasks list --status pending}
    \item \texttt{calendar list}
    \item \texttt{calendar today --timezone America/New\_York}
    \item \texttt{file create --path '/ops/release-handoff.txt' --content 'Boston refreshed release handoff notes.'}
    \item \texttt{tasks complete --title 'Existing Boston release follow-up'}
    \item \texttt{tasks add --title 'Boston release follow-through' --priority high --due 2026-03-16}
    \item \texttt{calendar add-event --title 'Boston release sync' --start 2026-03-19T15:00}
\end{enumerate}
\tcblower
\small
\textbf{Evaluation outcome.} This rollout passes only if the stale object is actually retired and both replacement artifacts are created. It fails when the model notices the conflict but leaves the outdated task in place, adds the new task without cleanup, or stops after a partial repair and behaves as if the workflow were complete.
\end{tcolorbox}

\begin{tcolorbox}[colback=blue!4,colframe=blue!40!black,title={{\small Example L: Completion Gap (\texttt{hard\_decision\_workflow\_199})}},fonttitle=\bfseries\small,left=6pt,right=6pt,top=4pt,bottom=4pt]
\small\label{app:ex-completion-gap}
\textbf{Representative instance.} \texttt{hard\_decision\_workflow\_199} asks: ``Part of the Seattle release work is already handled, but the next step is the missing piece. Check what's there, read the release notes, refresh the handoff file, and leave the sync on the calendar in place and add only the missing next step on the board.'' \\[4pt]
\textbf{Initial state.} The override layer preloads the review note \texttt{/ops/release-review.txt} and an existing Seattle release sync on the calendar. What is missing is the board next step plus the refreshed handoff file. \\[4pt]
\textbf{Required correction.} Read the drafted release notes, preserve the existing Seattle sync, create the handoff file, and add only the missing high-priority task \texttt{Seattle release next step} on the board. \\[4pt]
\textbf{Representative actions.}
\begin{enumerate}[leftmargin=18pt, itemsep=2pt, topsep=2pt]
    \item \texttt{openclaw config get agent.model}
    \item \texttt{tasks list --status pending}
    \item \texttt{calendar list}
    \item \texttt{calendar today --timezone America/Los\_Angeles}
    \item \texttt{file read --path '/ops/release-review.txt'}
    \item \texttt{file create --path '/ops/release-handoff.txt' --content 'Seattle completion-gap release handoff notes.'}
    \item \texttt{calendar list --from 2026-03-10 --to 2026-03-10}
    \item \texttt{tasks add --title 'Seattle release next step' --priority high --due 2026-03-08}
\end{enumerate}
\tcblower
\small
\textbf{Evaluation outcome.} Evaluation checks exactly that structure: the handoff file must be created, the release sync must remain an already-existing artifact, no duplicate calendar event may be created, and the new task must appear. This is a real gap-completion example where the missing piece is the board task rather than the calendar sync.
\end{tcolorbox}

\begin{tcolorbox}[colback=orange!3,colframe=orange!40!black,title={{\small Example M: Branch Resolution (\texttt{hard\_decision\_workflow\_211})}},fonttitle=\bfseries\small,left=6pt,right=6pt,top=4pt,bottom=4pt]
\small\label{app:ex-branch-resolution}
\textbf{Representative instance.} \texttt{hard\_decision\_workflow\_211} asks: ``The Sydney ops setup already has part of the work in place. Please review the forecast, today's Australia/Sydney calendar, the task already on the board, and the recurring daily check schedule before working out whether this stays live or moves async, and send bob@example.com the update if it moves async.'' \\[4pt]
\textbf{Initial state.} The seeded state already contains \texttt{Sydney existing ops next-step task} and an active daily-check cron. The task instance resolves to the async branch, so the live review path should not be scheduled. \\[4pt]
\textbf{Required correction.} Inspect the Sydney forecast, calendar, board task, and cron state; decide that the workflow moves async; preserve the existing task and daily check; and send the backup-plan email to \texttt{bob@example.com} instead of creating a calendar review block. \\[4pt]
\textbf{Representative actions.}
\begin{enumerate}[leftmargin=18pt, itemsep=2pt, topsep=2pt]
    \item \texttt{weather forecast --location 'Sydney' --days 1}
    \item \texttt{calendar today --timezone Australia/Sydney}
    \item \texttt{openclaw cron list}
    \item \texttt{tasks list --status pending}
    \item Send the Sydney backup-plan email to \texttt{bob@example.com}.
\end{enumerate}
\tcblower
\small
\textbf{Evaluation outcome.} The evaluator checks the preserved task and cron state, the absence of duplicate board or cron artifacts, the email side effect, and the absence of a duplicate calendar event. This makes the task a concrete branch-resolution example instead of a generic live-vs-async summary.
\end{tcolorbox}

\begin{tcolorbox}[colback=green!2,colframe=green!30!black,title={{\small Example N: Already-Done Skip (\texttt{hard\_decision\_workflow\_251})}},fonttitle=\bfseries\small,left=6pt,right=6pt,top=4pt,bottom=4pt]
\small\label{app:ex-already-done-skip}
\textbf{Representative instance.} \texttt{hard\_decision\_workflow\_251} asks: ``The New York release setup should already be in place. Please verify the current model, the next-step task on the board, and the sync on the calendar before you send finance@example.com a short recap and leave it alone if it is already right.'' \\[4pt]
\textbf{Initial state.} Both key artifacts are already present through overrides: \texttt{New York existing release next step} on the board and \texttt{New York existing release sync} on the calendar. The only remaining obligation is to verify the state and send the finance recap. \\[4pt]
\textbf{Required correction.} Confirm the current model, inspect the board and calendar, recognize that no repair is needed, and send the requested recap to \texttt{finance@example.com} without creating any new task or event. \\[4pt]
\textbf{Representative actions.}
\begin{enumerate}[leftmargin=18pt, itemsep=2pt, topsep=2pt]
    \item \texttt{openclaw config get agent.model}
    \item \texttt{tasks list --status pending}
    \item \texttt{calendar list}
    \item \texttt{calendar today --timezone America/New\_York}
    \item Send the New York release recap email to \texttt{finance@example.com}.
\end{enumerate}
\tcblower
\small
\textbf{Evaluation outcome.} This task passes only when the model resists unnecessary mutation. The evaluator checks the model setting, the existing task and sync, the recap email, and the absence of duplicate task or calendar creations. It is therefore a real recognition-and-restraint task, not an empty no-op.
\end{tcolorbox}

\begin{tcolorbox}[colback=red!6,colframe=red!50!black,title={{\small Example O: Wrong-State Replacement (\texttt{hard\_decision\_workflow\_261})}},fonttitle=\bfseries\small,left=6pt,right=6pt,top=4pt,bottom=4pt]
\small\label{app:ex-wrong-state-replacement}
\textbf{Representative instance.} \texttt{hard\_decision\_workflow\_261} asks: ``Something in the Seattle release setup was staged wrong. Check whether the next step is stale, replace the stale next step with the correct version, and leave the next step and sync correctly set.'' \\[4pt]
\textbf{Initial state.} Under \texttt{hard\_wrong\_task}, the workflow starts with \texttt{Existing Seattle release follow-up}, a low-priority stale task that should not survive in the final state. The replacement task and replacement sync are both still missing. \\[4pt]
\textbf{Required correction.} Inspect the seeded Seattle task, determine that it is stale, complete it, create the corrected task \texttt{Seattle release replacement next step}, and add the matching \texttt{Seattle replacement release sync} event. \\[4pt]
\textbf{Representative actions.}
\begin{enumerate}[leftmargin=18pt, itemsep=2pt, topsep=2pt]
    \item \texttt{tasks list --status pending}
    \item \texttt{calendar list}
    \item \texttt{calendar today --timezone America/Los\_Angeles}
    \item \texttt{tasks complete --title 'Existing Seattle release follow-up'}
    \item \texttt{tasks add --title 'Seattle release replacement next step' --priority high --due 2026-03-09}
    \item \texttt{calendar add-event --title 'Seattle replacement release sync' --start 2026-03-10T13:00}
\end{enumerate}
\tcblower
\small
\textbf{Evaluation outcome.} Strict evaluation requires both retirement of the stale task and creation of the corrected replacement state. Adding the replacement without cleanup, or cleaning up without recreating the follow-through, still fails. That is why this slice remains one of the hardest state-conflict settings in the benchmark.
\end{tcolorbox}

\begin{tcolorbox}[colback=cyan!2,colframe=cyan!35!black,title={{\small Example P: Interrupted Workflow Resume (\texttt{hard\_decision\_workflow\_307})}},fonttitle=\bfseries\small,left=6pt,right=6pt,top=4pt,bottom=4pt]
\small\label{app:ex-interrupted-workflow-resume}
\textbf{Representative instance.} \texttt{hard\_decision\_workflow\_307} asks: ``The New York release work already has pieces in place. Check the current model, the next step on the board, and the sync on the calendar, finish the missing handoff file, send manager@example.com a recap, and leave the existing setup alone.'' \\[4pt]
\textbf{Initial state.} The seeded release state already contains \texttt{New York existing release next step}, \texttt{New York existing release sync}, and drafted notes in \texttt{/ops/release-review.txt}. The only missing artifact is the handoff file, after which a recap email is required. \\[4pt]
\textbf{Required correction.} Resume from the partially completed New York release setup, read the existing review notes, create the missing handoff file, preserve the staged task and sync, and send the recap to \texttt{manager@example.com}. \\[4pt]
\textbf{Representative actions.}
\begin{enumerate}[leftmargin=18pt, itemsep=2pt, topsep=2pt]
    \item \texttt{openclaw config get agent.model}
    \item \texttt{tasks list --status pending}
    \item \texttt{tasks search --query 'New York release'}
    \item \texttt{calendar list}
    \item \texttt{calendar today --timezone America/New\_York}
    \item \texttt{file read --path '/ops/release-review.txt'}
    \item \texttt{file create --path '/ops/release-handoff.txt' --content 'New York resumed release handoff notes.'}
    \item Send the New York resume recap email to \texttt{manager@example.com}.
\end{enumerate}
\tcblower
\small
\textbf{Evaluation outcome.} The evaluator accepts only resume-style closure: the task and sync must remain existing artifacts, no duplicates may be created, the handoff file must appear, and the recap email must be sent. Restarting the workflow from scratch or mutating already-correct state still fails.
\end{tcolorbox}

\begin{tcolorbox}
[colback=purple!2,colframe=purple!40!black,title={{\small Example Q: Contradictory Source Resolution (\texttt{hard\_decision\_workflow\_337})}},fonttitle=\bfseries\small,left=6pt,right=6pt,top=4pt,bottom=4pt]
\small\label{app:ex-contradictory-source-resolution}
\textbf{Representative instance.} \texttt{hard\_decision\_workflow\_337} asks: ``Please reconcile the latest review email note, the forecast, and today's Europe/London calendar for London. Leave the next step on the board, keep the recurring daily check scheduled, and either put the review on the calendar or send bob@example.com the async update.'' \\ [3pt]
\textbf{Initial state.} This London instance begins without seeded overrides, so all action depends on reconciling the latest review email, the forecast, and the current London calendar. The correct branch here is async: the task and recurring check should be created, but no review event should be added. \\ [3pt]
\textbf{Required correction.} Read the latest review email note, inspect forecast and calendar evidence, create \texttt{London contradictory-source next step}, add the recurring contradiction-check cron, and send the async update to \texttt{bob@example.com} while avoiding a calendar review block. \\ [2pt]
\textbf{Representative actions.}
\begin{enumerate}[leftmargin=18pt, itemsep=2pt, topsep=2pt]
    \item \texttt{email search --query 'review'}
    \item \texttt{email read --id email\_seed\_5}
    \item \texttt{weather forecast --location 'London' --days 1}
    \item \texttt{calendar today --timezone Europe/London}
    \item \texttt{calendar list --from 2026-03-01 --to 2026-03-01}
    \item \texttt{tasks list --status pending}
    \item \texttt{tasks add --title 'London contradictory-source next step' --priority high --due 2026-03-11}
    \item \texttt{openclaw cron add --name contradictory-source-01 --cron '0 18 * * 1-5' --message 'Run London daily contradiction check'}
    \item Send the London async update email to \texttt{bob@example.com}.
\end{enumerate}
\tcblower
\small
\textbf{Evaluation outcome.} The evaluator checks exactly those branch-specific side effects: task created, cron created, email sent, and no duplicate calendar event. This makes the example a genuine contradictory-evidence resolution task rather than a generic multi-source lookup problem.
\end{tcolorbox}

\end{document}